\documentclass{article}

\usepackage[preprint,nonatbib]{neurips_2022}
\usepackage[square,sort,comma,numbers]{natbib}
\makeatletter
\usepackage{color}
\usepackage[utf8]{inputenc} % allow utf-8 input
\usepackage[T1]{fontenc}    % use 8-bit T1 fonts
\usepackage{hyperref}       % hyperlinks
\usepackage{url}            % simple URL typesetting
\usepackage{booktabs}       % professional-quality tables
\usepackage{amsfonts}       % blackboard math symbols
\usepackage{nicefrac}       % compact symbols for 1/2, etc.
\usepackage{microtype}      % microtypography
\usepackage{xcolor}         % colors
\usepackage{amssymb}
\usepackage{graphicx}
\usepackage{amsmath}
\usepackage{float}

\usepackage{subfig}
\usepackage{multirow}
\usepackage{subfloat}
\usepackage{parskip}
\usepackage{authblk}
\usepackage{wrapfig,lipsum,booktabs}
\usepackage{enumitem,kantlipsum}
\usepackage[linesnumbered, ruled]{algorithm2e}  
\SetKwRepeat{Do}{do}{while}%  

\title{Single-Stage Open-world Instance Segmentation with Cross-task Consistency Regularization} 
\author[a,d]{Xizhe Xue \thanks{This work was done when Xizhe Xue visited National University of Singapore. Email: xuexizhe@mail.nwpu.edu.cn}}
\author[b]{Dongdong Yu}
\author[c]{Lingqiao Liu}
\author[b]{Yu Liu}
\author[d]{Satoshi Tsutsui}
\author[a]{Ying Li\thanks{Corresponding author}}
\author[b]{Zehuan Yuan}
\author[b]{Ping Song}
\author[d]{Mike Zheng Shou}
\affil[a]{Northwestern Polytechnical University}
\affil[b]{ByteDance Inc.}
\affil[c]{University of Adelaide}
\affil[d]{Show Lab, National University of Singapore}

\begin{document}
\maketitle

\begin{abstract}

Open-World Instance Segmentation (OWIS) is an emerging research topic that aims to segment class-agnostic object instances from images. The mainstream approaches use a two-stage segmentation framework, which first locates the candidate object bounding boxes and then performs instance segmentation. In this work, we instead promote a single-stage framework for OWIS. We argue that the end-to-end training process in the single-stage framework can be more convenient for directly regularizing the localization of class-agnostic object pixels. Based on the single-stage instance segmentation framework, we propose a regularization model to predict foreground pixels and use its relation to instance segmentation to construct a cross-task consistency loss. We show that such a consistency loss could alleviate the problem of incomplete instance annotation -- a common problem in the existing OWIS datasets. We also show that the proposed loss lends itself to an effective solution to semi-supervised OWIS that could be considered an extreme case that all object annotations are absent for some images. Our extensive experiments demonstrate that the proposed method achieves impressive results in both fully-supervised and semi-supervised settings. Compared to SOTA methods, the proposed method significantly improves the $AP_{100}$ score by 4.75\% in UVO$\rightarrow$UVO setting and 4.05\% in COCO$\rightarrow$UVO setting. In the case of semi-supervised learning, our model learned with only 30\% labeled data, even outperforms its fully-supervised counterpart with 50\% labeled data. The code will be released soon at: \href{https://github.com/showlab/SOIS.git}{https://github.com/showlab/SOIS}.

\end{abstract}
\section{Introduction}

Traditional instance segmentation~\cite{coco,cityscapes} methods often assume that objects in images can be categorized into a finite set of predefined classes (i.e., \textit{closed-world}). Such an assumption, however, can be easily violated in many real-world applications, where models will encounter many new object classes that never appeared in the training data. Therefore, researchers recently attempted to tackle the problem of \textbf{Open-World Instance Segmentation (OWIS)}~\cite{uvo},  which targets class-agnostic segmentation of all objects in the image. 

Prior to this paper, most existing methods for OWIS are of two-stage~\cite{ggn,ldet}, which detects bounding boxes of objects and then segments them. Despite their promising performances, such a paradigm cannot handle and recover if object bounding boxes are not detected. In contrast, a single-stage approach called Mask2Former~\cite{mask2former} has recently been introduced, yet only for closed-world instance segmentation. By extending it to open-world, we are \textbf{the first to} develop a novel \textbf{detection-free Single-stage Open-world Instance Segmentation method, dubbed as SOIS}. 
 
%In fact, two-stage methods sometimes outperform simple Mask2Former baselines (Sec. \ref{sec:not-just-mask2form}), so we address challenges unique to open-world scenarios as follows. 
%Our method is motivated by the observation that the instance annotation in the existing datasets is incomplete. Unlike in closed-world instance segmentation, where the object categories have been clearly defined, instance definition in OWIS is much more ambiguous and harder for annotators to follow. Inevitably, the instance annotation could become inconsistent across images. 

\begin{figure*}[]
	\begin{center}
		\includegraphics[width=0.95\linewidth]{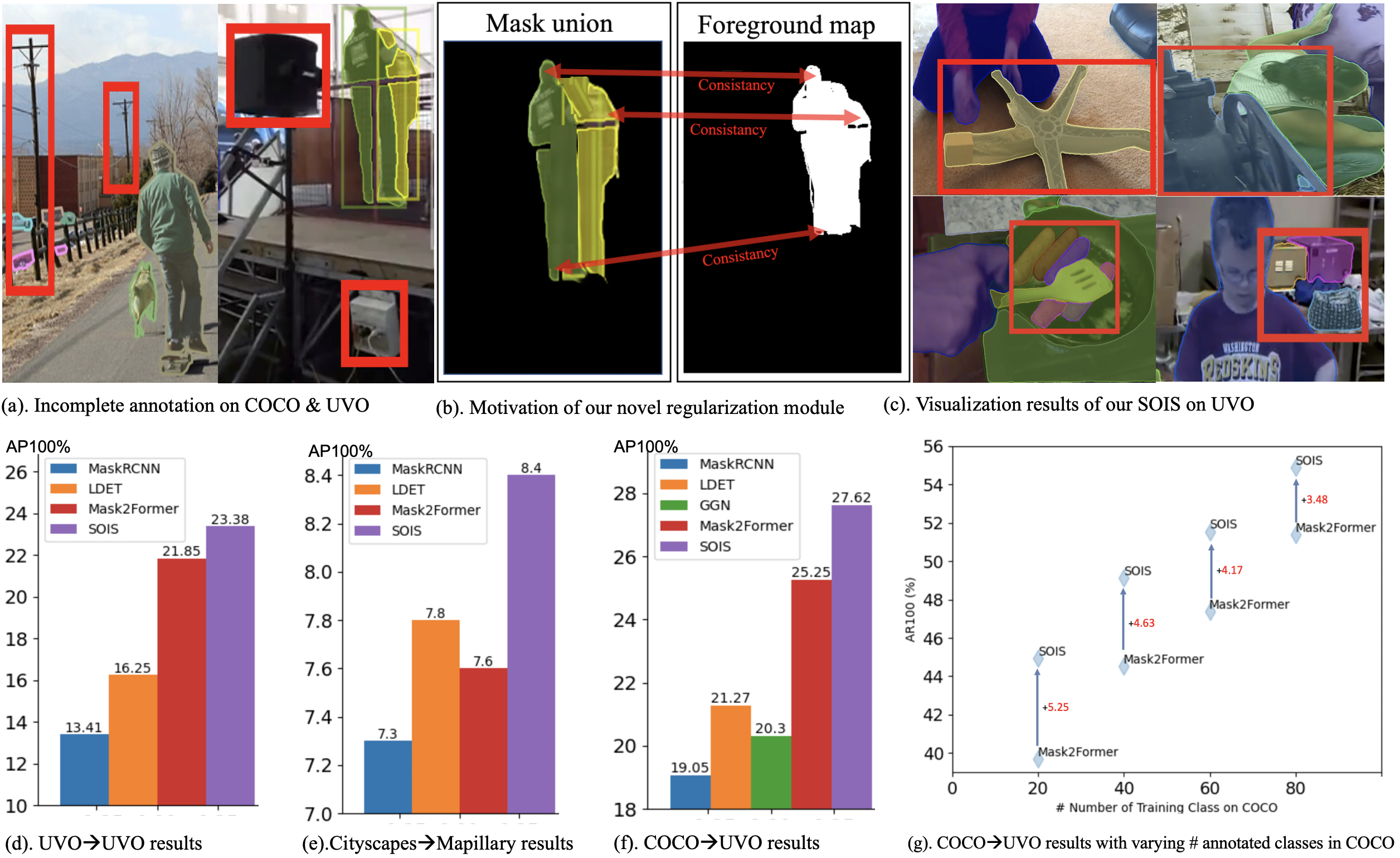}
	\end{center}
	\caption{ \textbf{(a).} Instances missing annotations in COCO and UVO datasets. The regions in \textcolor{red}{red boxes} are mistakenly annotated as background. \textbf{(b).} Motivation of our novel reg module (The consistency relationship between instance mask and foreground map). \textbf{(c).} Visualization results of our SOIS on UVO dataset. Here, the proposed SOIS is trained on COCO dataset and tested on UVO dataset. Our methods correctly segments many objects that are not labeled in COCO. \textbf{(d - f).} The $AP_{100}$\% of our SOIS \textit{vs.}SOTA methods on COCO$\rightarrow$UVO, Cityscpes$\rightarrow$Mapillary, COCO$\rightarrow$UVO. 
	\textbf{(g).} The $AR_{100}$\% of our SOIS \textit{vs.} baseline Mask2Former on COCO.
	%Part COCO (with 20 classes, 40 classes, 60 classes, 80 classes of object annotation)$\rightarrow$UVO. 
	From right to left, with the total number of classes decreases (i.e. more instance annotations missed), the gain of our SOIS over baseline becomes larger, thanks to the capability of our model to handle incomplete annotations.}
	\label{fig:fig0}
\end{figure*}

Note that our work is not just a straightforward adaptation of Mask2Former from close-world to open-world. This is because unlike closed-world segmentation, where the object categories can be clearly defined before annotation, the open-world scenario makes it challenging for annotators to label all instances completely or ensure annotation consistency across different images because they cannot have a well-defined finite set of object categories. As shown in Figure \ref{fig:fig0}(a), annotators miss some instances.
\textbf{It still remains challenging that how to handle such incomplete annotations (i.e. some instances missed)}.  

Recent work LDET~\cite{ldet} addresses this problem by generating synthetic data with a plain background, but it based on a decoupled training strategy that can only be used in the two-stage method \textit{while our method is of single-stage}. Another work called GGN~\cite{ggn} handles such incomplete instance-level annotation issue by training a pairwise affinity predictor for generating pseudo labels. But training such an additional predictor is complicated and time-consuming. 

In contrast, \textit{our proposed SOIS method is end-to-end and simpler}. We address this incomplete annotation issue via a \textbf{novel regularization module}, which is simple yet effective. Specifically, it is convenient to concurrently predict not only (1) \textit{instance masks} but also a (2) \textit{foreground map}. Ideally, as shown in Figure \ref{fig:fig0}(b), the foreground region should be consistent with the union of all instance masks.
To penalize their inconsistency, we devise a \textit{cross-task consistency loss}, which can down-weight the adverse effects caused by incomplete annotation.
This is because when an instance is missed in annotation, as long as it is captured by both our predictions of instance masks and foreground map, the consistency loss would be low and hence encourage such prediction.
Experiments in Figure \ref{fig:fig0}(g) show that such consistency loss is effective even when annotations miss many instances.

%Such consistency between the foreground map and instance masks should always hold, regardless of missing instance annotations. Inspired by this observation, we design our regularization module to down-weight the adverse effects caused by incomplete annotation. Specifically, we first compute the union of all instance masks, which should be pixels falling into \textit{any} object instances. Then we create a prediction branch to generate the \textit{foreground region prediction}. We promote the inherent relation between the foreground prediction and the instance prediction by designing a regularization term named . We experimentally show that our consistency loss is particularly effective when annotations have many missing instances (Figure \ref{fig:fig0}(g)).

% We explicitly model such consistency by designing a regularization term named \textit{cross-task consistency loss} to penalize the differences between our predicted foreground region and the union of all predicted instance masks.
% We find that such regularization can down-weight the adverse effects to our correctly predicted instances that do not have ground truth annotation and .

So far, like most existing methods, we focus on the fully-supervised OWIS.
In this paper, we further extend OWIS to the semi-supervised setting, where some training images do not have any annotations at all.
This is of great interest because annotating segmentation map is very costly.
Notably, \textbf{our proposed regularization module can also benefit semi-supervised OWIS} -- consider an unlabeled image as an extreme case of incomplete annotation where all of the instance annotations are missed.
Specifically, we perform semi-supervised OWIS by first warming up the network on the labeled set and then continuing training it with the cross-task consistency loss on the mixture of labeled and unlabeled images. 

\textbf{Contributions}. In a nutshell, our main contributions could be summarized as:
\vspace{-0.5em}
\begin{enumerate}
  \setlength{\parskip}{0cm} % 段落間
  \setlength{\itemsep}{0.1em} % 項目間
  \item We propose a Singe-stage Open-world Instance Segmentation (SOIS) for the first time while most OWIS methods are of two-stage.
  \item We propose a novel cross-task consistency loss that mitigate the issue of incomplete mask annotations. 
  \item We further extend the proposed method into a semi-supervised OWIS model, which effectively makes use of the unlabeled images to help the OWIS model training . 
  \item Our extensive experiments demonstrate that the proposed method reaches the leading OWIS performance in the fully-supervised learning. (Figure~\ref{fig:fig0}(d-f)), and that our semi-supervised extension can achieve remarkable performance with a much smaller amount of labeled data.
\end{enumerate}

\section{Related Work}
%\subsection{Closed-world instance segmentation} 
\paragraph{Closed-world instance segmentation (CWIS)} ~\cite{maskrcnn,blendmask,center-mask,yolact,solo} requires the approaches to assign a class label and instance ID to every pixel. Two-stage CWIS approaches, such as MaskRCNN, always include a bounding box estimation branch and a FCN-based mask segmentation branch, working in a 'detect-then-segment' way. To improve efficiency, one-stage methods such as CenterMask~\cite{center-mask}, YOLACT~\cite{yolact} and BlendMask~\cite{blendmask} have been proposed, which remove the proposal generation and feature grouping process. To further free the CWIS from the local box detection, Wang et.al ~\cite{solo} proposed SOLO and obtained on par results to the above methods. In recent years, the methods ~\cite{instances,solq}, following DETR~\cite{detr}, consider the instance segmentation task as an ensemble prediction problem. In addition, Cheng et al. proposed an universal segmentation framework MaskFormer~\cite{maskformer} and its upgrade version Mask2Former~\cite{mask2former}, which even outperforms the state-of-the-art architectures specifically designed for the CWIS task.

Notably, two-stage method CenterMask preserves pixel alignment and separates the object simultaneously by integrating the local and global branch. Although introducing the global information in this way helps improve the mask quality in CWIS, it can not handle the open-world task very well. Because CenterMask multiplies the local shape and the cropped saliency map to form the final mask for each instance. There is no separate loss for the local shape and global saliency. When such method faces the incomplete annotations in OWIS tasks, the generated mask predictions corresponding to the unlabeled instances would still be punished during training, making it difficult to discover novel object at inference. The efficient way to jointly take advantages of global and local information in OWIS tasks deserves to be explored.

%%%Space issues, this section does not seem to be relevant to our content either, lightning
% %\subsection{Open-world detection and entity segmentation} 
% \paragraph{Open-world detection and entity segmentation} 
% The open-world object detection (OWOD)~\cite{OLN,owdetr,owp} is first proposed in~\cite{owod}. It aims to detect known categories of objects and identify unknown objects, while requiring the method to have the ability to lean new classes. To address this problem, Joseph et al.~\cite{owod} constructed ORE, which comprises three dedicated components namely contrastive clustering module, unknown-aware proposal network and the energy-based unknown identification module. Another method OW-DETR~\cite{owdetr} employs the attention-driven pseudo-labeling, novelty classification, and objectness scoring jointly to meet OWOD challenges. Open-world entity segmentation task (OWES)~\cite{owes} aims to segment all entities (instances and stuff) in an image without predicting their categories. To solve this problem, Qi et al.~\cite{owes} proposed a center-based entity segmentation framework to improve the quality of the entity mask.

\paragraph{Open-world instance segmentation} OWIS task~\cite{uvo} here focuses on the following aspects: (1) All instances (without stuff) have to be segmented; (2) Class-agnostic pixel-level results should be predicted with only instance ID and incremental learning ability is unnecessary. Several OWIS works have recently been developed. Yu et al.~\cite{winner} proposed a two-stage segmentation algorithm, which decoupled the segmentation and detection modules during training and testing. This algorithm achieves competitive results on the UVO dataset thanks to the abundant training data and the introduction of effective modules such as cascade RPN~\cite{vu2019cascade}, SimOTA~\cite{yolox}, etc. Another work named LDET~\cite{ldet} attempts to solve the instance-level incomplete annotation problem. Specifically, LDET first generates the background of the synthesized image by taking a small piece of background in the original image and enlarging it to the same size as the original image.  The instance is then matted to the foreground of the synthesized image. The synthesized data is used only to train the mask prediction branch, and the rest of the branches are still trained with the original data.  Meanwhile, Wang et al. proposed GGN~\cite{ggn}, an algorithm that combines top-down and bottom-up segmentation ideas to improve prediction accuracy by generating high-quality pseudo-labels. Specifically, a Pairwise Affinity (PA) predictor is trained first and a grouping module is used to extract and rank segments from predicted PA to generate pseudo-labels, which would be fused with groundtruth to train the segmentation model.

\section{Methodology}
In this section, we first define the OWIS problem with both fully and semi-supervised learning.  Then the architecture of our SOIS and the proposed cross-task consistency loss are introduced in Section 3.3 and 3.4, respectively. Finally, Section 3.4 and 3.5 show how to optimize the SOIS in fully and semi-supervised way, respectively.

\subsection{Problem Definition of OWIS} The open-world instance segmentation (OWIS) aims to segment all the object instances (things) of any class including those that did not appear in the training phase. Technically, OWIS is a task to produce a set of binary masks, where each mask corresponds to a class-agnostic instance. The pixel value of $1$ in the mask indicates a part of an object instance while $0$ indicates not.

% In fully-supervised OWIS setting, all the training images have been annotated, while semi-supervised OWIS models could make use of the unlabeled images. 

% Following the recent OWIS works, we design our single-stage network architecture with cross-task consistency regularization to pursue both efficiency and effectiveness under the fully supervised setting. Moreover, the proposed framework enables us to leverage the unlabeled images in a semi-supervised way for achieving comparable performance with lower human labeling cost.

\subsection{Model Architecture}

\begin{figure*}[]
	\begin{center}
		\includegraphics[width=0.95\linewidth]{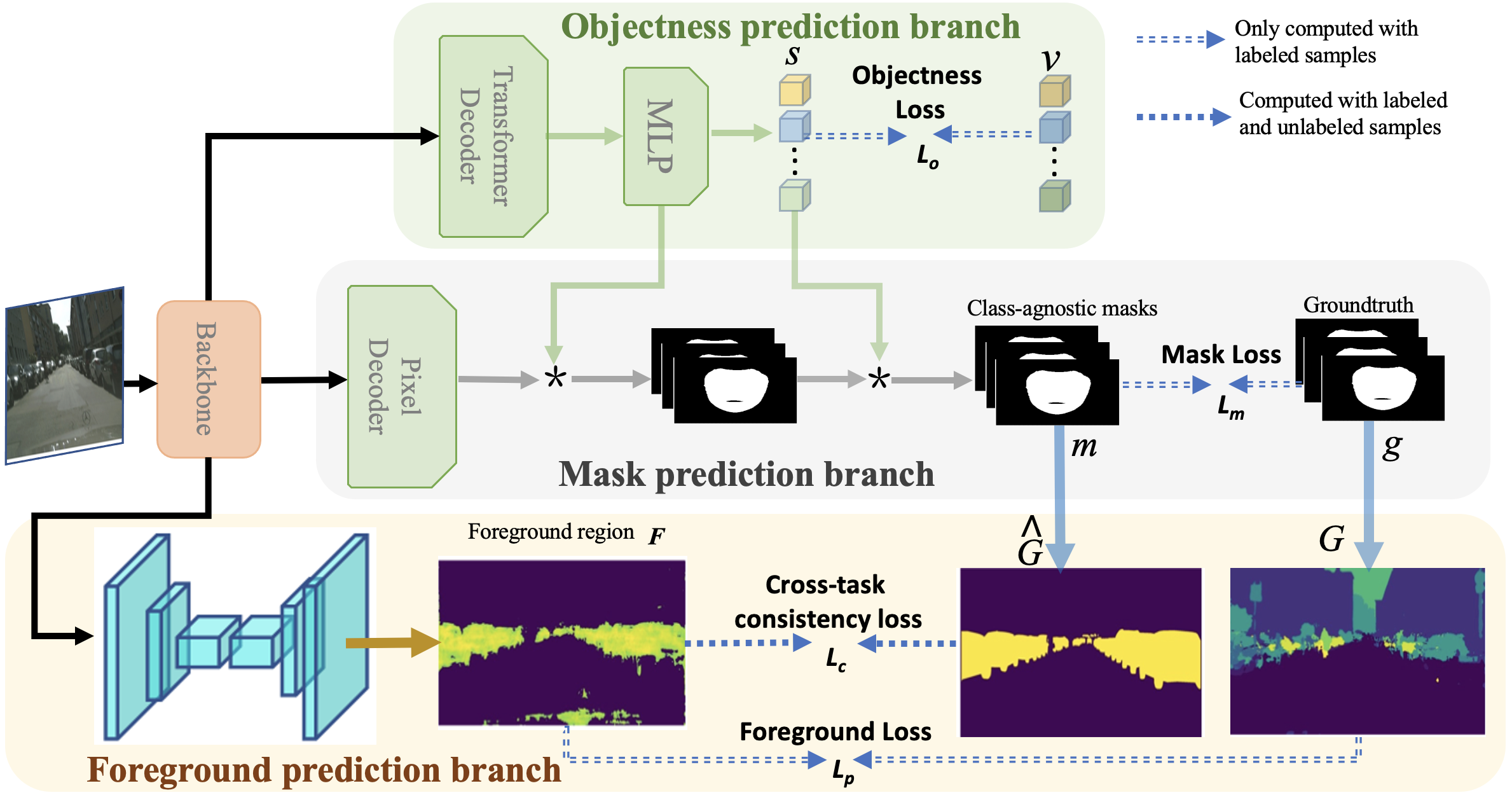}
	\end{center}
    \raggedbottom
	\caption{ Overall framework of the proposed SOIS. The mask prediction branch generates the predicted masks, while the objectness prediction branch computes the objectness score for each mask. The foreground prediction branch segments a foreground region to guide the optimization of other two branches.} 
\label{fig:fig1}
\end{figure*}

Our proposed SOIS framework consists of three branches to alleviate the incomplete annotating, as shown in Figure~\ref{fig:fig1}. Basically, we follow the design of one-stage Mask2Former~\cite{mask2former}. The \textbf{objectness prediction branch} estimates the weighting score for each mask by applying a sequential Transformer decoder and MLP. The \textbf{mask prediction branch} predicts the binary mask for each instance. It first generates $N$ binary masks with $N$ ideally larger than the actual instance number $K_i$. Each mask is multiplied by a weighting score with a value between $0$ and $1$, indicating if a mask should be selected as an instance mask.  This process generates the mask in an end to end way, which avoids to miss the instance because of poor detection bounding boxes and meanwhile reduces the redundant segmentation cost for each proposal. We refer to \cite{maskformer,mask2former} for more details including the training procedure.

The \textbf{foreground prediction branch} is a light-weight fully convolutional network to estimate the foreground regions that belong to any object instance.The more detailed design of the foreground prediction branch is in the Appendix. This guides the training of the mask branch through our cross-task consistency loss proposed in the following Sec.~\ref{sec:consistency_loss}. Once training is done, we discard this branch and only use the objectness and mask prediction branch at inference time. Therefore, we would not introduce any additional parameter or computational redundancy, which benefits the running efficiency.

\subsection{Learning with the cross-task consistency regularization}\label{sec:consistency_loss}

A critical limitation of the OWIS is the never-perfect annotations due to the difficulties in annotating class-agnostic object instances. Towards alleviating this issue, we propose a regularization to provide extra supervision to guide the OWIS model training under incomplete annotations.

We construct a branch to predict the foreground regions that belong to any of the object instance. Formally we create the the foreground annotation $G(x,y)$ calculated by 
\begin{eqnarray}
{G}(x,y)= \begin{cases}
 \text{ 0, }\: \; \text{if} \: \; \sum_{i= 1}^{K}{g^{i}}(x,y)==0  \\ 
 \text{ 1,}\: \; \text{otherwise,}
\end{cases}
\label{eq:foreground_instance_relation}
\end{eqnarray}where $g^i(x,y)$ is one of the $K$ annotated object instances for the current image and the union of $g^i$ defines the foreground object regions. Here $(x,y)$ denotes a coordinate of a pixel in the an image. We use $G(x,y)$ as labels to train the foreground prediction branch. 

Our consistency loss encourages the model outputs to have the relationship indicated in Eq \ref{eq:foreground_instance_relation}, which states that the the foreground prediction should be the union of instance predictions. To do so, we use the following equation as an estimate of the foreground from the instance prediction:
\begin{align}
    \hat{G}(x,y)= \Phi \left (\sum_{j= 1}^{K}{m^{j}}(x,y)\right ),
\end{align}where $m^j$ means the confidence of pixels in j-th predicted mask, and $\Phi$ represents the Sigmoid function. Then, let the foreground prediction from the foreground prediction branch be $F$, our cross-task consistency loss is to make $F$ and \(\hat{G}(x,y)\) consistent, which finally leads to the following loss function.

\begin{eqnarray}
L_{c}=\text{DICE}_{( \hat{G},F)}+\text{BCE}_{( \hat{G},F)},
\label{eq3}
\end{eqnarray}
where DICE and BCE denote the dice-coefficient loss~\cite{dice} and binary cross-entropy loss, respectively. 

\begin{figure}
\includegraphics[width=0.95\linewidth,]{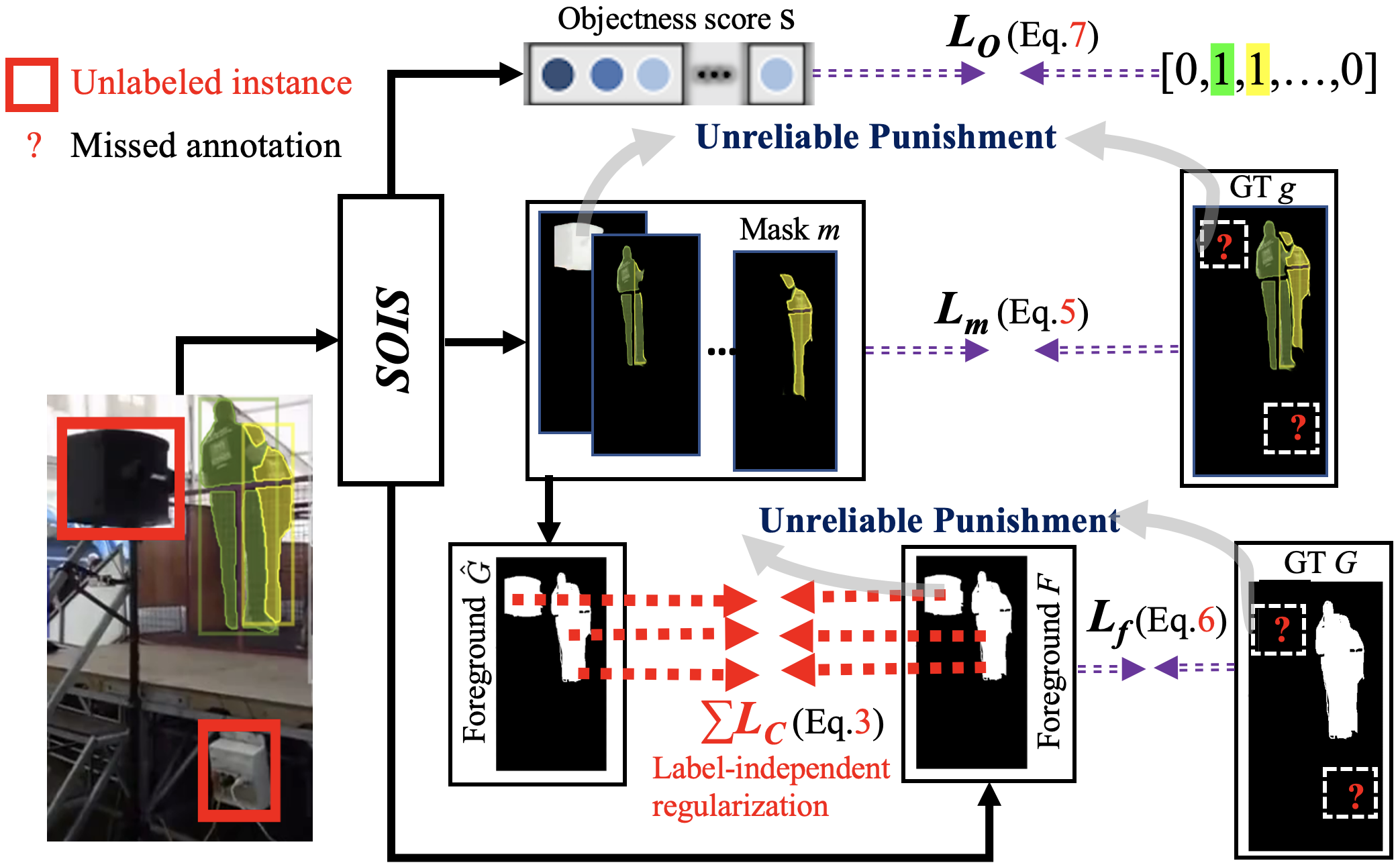}
\caption{Working principle of consistency loss.}
\label{fig:consistancy_loss}
\end{figure}
Consistency loss enjoys the following appealing properties. It is self-calibrated and independent with the incompleteness level of labels. As shown in Figure~\ref{fig:consistancy_loss}, for a instance mistakenly annotated as background, but the foreground prediction branch and mask prediction branch both correctly find it, the model would be punished through mask loss and foreground loss. However, the consistency loss think this prediction is correct. In this way, consistency loss down-weights the adverse effects caused by other unreliable segmentation loss. The mitigation and the compensation factor synergize to relieve the overwhelming punishments on unlabeled instances.

% For an instance mistakenly annotated as background, but the foreground prediction branch and mask prediction branch both correctly detect it, the mask loss, objectness loss and foreground loss are increased, while the consistency loss is low. In this way, consistency loss down-weights the adverse effects caused by other unreliable segmentation loss. The mitigation and the compensation factor synergize to relieve the overwhelming punishments on unlabeled instances.

%\paragraph{Motivation}In the open world setting, positive samples of the training data always belong to a specific domain, while instances appearing in the test set may be out-of-distribution. It is difficult to segment unseen or novel objects, while much easier to identify all the foreground pixels (belonging to any object instance). Therefore, we construct an auxiliary foreground prediction task to guide the optimization of the challenging OWIS task. Apart from the per-pixel information, the obtained intermediate foreground map also contains the potential mutual co-existence relationship between objects, which serves as another effective clue to help jointly infer the state of unknown objects based on the states of some easily segmented objects.

\subsection{Fully-supervised learning}
% \paragraph{Fully-supervised learning process} We first generate the foreground groundtruth $G_{gt}$ and compute the global mask prediction $G_{mask}$ from the individual mask predictions according to Equation (\ref{eq1}):
% \begin{eqnarray}
% G_{gt}(x,y)= \begin{cases}
%  \text{ 0, }\: \; \text{if} \: \; \sum_{i= 1}^{N}{g^{i}}(x,y)==0  \\ 
%  \text{ 1,}\: \; \text{otherwise,}
% \end{cases}
% G_{mask}(x,y)= \Phi \left (\sum_{j= 1}^{N}{m^{j}}(x,y)\right )
% \label{eq1}
% \end{eqnarray}
%%%the code to generate G_mask is             
%sum_mask=torch.cat(sum_mask, dim=0) 
%sum_mask=F.sigmoid(torch.max(sum_mask,dim=0).values.float())
% Meanwhile, a foreground prediction branch is introduced to perform binary segmentation, as present in Equation (\ref{eq2}).
% \begin{eqnarray}
% H=\pounds (\delta ( \varphi ( \delta  (x_i;\theta ))))
% \label{eq2}
% \end{eqnarray}
% where multi-scale feature of sample $x_i$ is utilized to optimize the model parameter $\theta$. $\pounds (.)$ denotes a sequential point-wise convolution and Sigmod function, $ \delta(.)$ and $ \varphi(.)$ denote the $gOctaveCBR$ block and $PallMS$ block respectively, as described in ~\cite{gao2020highly}.

The overall fully-supervised optimization of the proposed SOIS is carried out by minimizing the following joint loss formulation $L_f$,
\begin{eqnarray}
\label{eq4}
&& L_f=\alpha L_m+\beta L_p+\gamma L_c +\omega L_o, \\
\textrm{where}
&& L_m=\text{BCE}_{(m,g)}+\text{DICE}_{(m,g)}, \\
&& L_p=\text{BCE}_{(F,G)}+\text{DICE}_{(F,G)},\\
&& L_o=\text{BCE}_{(s,v)},
\end{eqnarray}
where $L_m$, $L_p$ and $L_o$ denote the loss terms for mask prediction, foreground prediction, and objectness scoring, respectively. $\alpha$, $\beta$, $\gamma$ and $\omega$ are the weights of the corresponding losses. $m$ and $g$ represent the predicted masks and corresponding groundtruth, respectively. $F$ and $G$ is the foreground prediction result and the generated foreground groundtruth, while the estimated objectness score is denoted with $s$. $v$ is a set of binary values that indicate whether each mask is an instance. Before computing the $L_m$, matching between the set of predicted masks and groundtruth has been done via the bipartite matching algorithm defined in \cite{mask2former}.

\subsection{Extension to semi-supervised learning}
%Pixel-level annotations for OWIS are expensive, because all objects in the image are required to be densely annotated. Therefore, it is necessary and meaningful to explore how to make full use of unlabeled data in the open-world setting. Semi-supervised learning is an important step toward reducing expensive laborious human supervision. Since the proposed alignment loss does not require additional annotation information, it allows the SOIS framework to learn in the semi-supervised setting.  The results of the easy-to-learn foreground segmentation task are utilized to guide the optimization of primary OWIS task.

Due to the ambiguity of the instance definition in OWIS, it is much harder for the annotators to follow the annotation instruction, and this could make the annotations for OWIS expensive. It is desirable if we can use unlabeled data to help train OWIS models. In this regard, our proposed cross-task consistency loss only requires the outputs of both predictors to have a consistent relationship indicated in \ref{eq:foreground_instance_relation}, and does not always need ground truth annotations. Thus, we apply this loss to unlabeled data, which becomes semi-supervised learning. Specifically, the easier-to-learn foreground prediction branch is able to learn well through a few labeled images in the warm-up stage.  Then the resulted foreground map can serve as a constraint to optimize the open-world mask predictions with the help of our cross-task consistency loss, when the labels do not exist. In this way, our Semi-SOIS achieves a good trade-off between the annotation cost and model accuracy.

\textbf{Semi-supervised learning process.} Given a labeled set $D_l=\left \{(x_i,y_i)\right \}_{i=1}^{N_l}$ and an unlabeled set $D_u=\left \{x_i\right \}_{i=1}^{N_u}$, our goal is to train an OWIS model by leveraging both a large amount of unlabeled data and a smaller set of labeled data. Specifically, we initially use $D_l$ to train the SOIS as a warm-up stage, giving a good initialization for the model. We then jointly train the OWIS model on the both labeled and unlabeled data. For the labeled data, we employ the loss function defined in Eq \ref{eq4}. For the unlabeled data, we apply only the cross-task consistency loss $L_{c}$.

\section{Experiments}
For demonstrating the effectiveness of our proposed SOIS, we compared it with other fully-supervised methods through intra-dataset and cross-dataset evaluations. We also performed ablation studies in these two settings to show the effect of each component. Moreover, we apply the proposed cross-task consistency loss for semi-supervised learning and test our method on the UVO validation set.

\subsection{Implementation details and evaluation metrics}
\paragraph{Implementation details} Detectron2~\cite{detectron2} is used to implement the proposed SOIS framework, multi-scale feature maps are extracted from the ResNet-50~\cite{he2016deep} or Swin Transformer~\cite{liu2021Swin} model pre-trained on ImageNet~\cite{deng2009imagenet}. Our transformer encoder-decoder design follows the same architecture as in Mask2Former~\cite{mask2former}. The number of object queries $M$ is set to 100. Both the ResNet and Swin backbones use an initial learning rate of 0.0001 and a weight decay of 0.05. A simple data augmentation method, Cutout~\cite{cutout}, is applied to the training data. All the experiments have been done on 8 NVIDIA V100 GPU cards with 32G memory.

\paragraph{Pseudo-labeling for COCO train set} Pseudo-labeling is a common way to handle incomplete annotations. To explore the compatibility of our method and the pseudo-labeling operation, we employ a simple strategy to generate pseudo-labels for unannotated instances in the COCO train set~\cite{coco} in our experiments. Specifically, we follow a typical self-training framework, introducing the teacher model and student model framework to generate pseudo-labels. These two models have the same architecture, as shown in Figure~\ref{fig:fig1}, but are different in model weights. The weights of the student model are optimized by the common back-propagation, while the weight of the teacher model is updated by computing the exponential moving averages (EMA) of the student model. During training, the image $i$ is first fed into the teacher model to generate some mask predictions. The prediction whose confidence is higher than a certain value would be taken as a pseudo-proposal. The state $S_{ij}$ of the pseudo-proposal $p_{ij}$ is determined according to Equation (\ref{eq9}).
\vspace{+0.2cm}
\begin{eqnarray}
S_{ij}= \begin{cases}
 \text{ True, }\: \; \text{if} \: \; \text{argmax}(\varphi(p_{ij},g_i)) \leqslant \varepsilon,  \\ 
 \text{ False,}\: \; \text{otherwise,}
\end{cases}
\label{eq9}
\end{eqnarray}
\vspace{+0.2cm}
in which $g_i$ means any ground truth instance in the image $i$. $\varphi$ denotes the IOU calculating function, and $\varepsilon$ is a threshold to further filter the unreliable pseudo-proposals. Finally, pseudo-proposals with states $True$ would be considered as reliable pseudo-labels. Here, the confidence and IOU threshold $\varepsilon$ for selecting pseudo-labels are set to 0.8 and 0.2, respectively. Then, we jointly use the ground truth and the pseudo-labels to form the training data annotations. If a region is identified as belonging to an instance in the pseudo-label, it will be considered as a positive sample during training.
%For image $i$, the corresponding annotation $G_i=p_i||g_i$, where $||$ means pixel-by-pixel $or$ operation.

\begin{table*}[t]
\renewcommand\arraystretch{1.0}
\setlength\tabcolsep{4pt}
    \centering
    \vspace{-0.2cm}
    \caption{Results of \textbf{UVO-train $\rightarrow$ UVO-val} intra-dataset evaluation.}  
    \vspace{-0.2cm}
    \label{table1}
    \resizebox{11cm}{!}{
    \begin{tabular}{l|l|ccccccc}
    \toprule
    Metric & Backbone  & AP$_{100}(\%)$ & AP$_{s}$(\%)  & AP$_{m}$(\%)  & AP$_{l}$(\%)  & AR$_{100}$(\%)  &AR$_{10}$(\%) \\
    %\midrule
    \hline
    MaskRCNN     & R-50    & 13.41  & 4.91  &  12.33  & 17.45  & 22.77  & 20.01  \\
    LDET         & R-50    & 16.25  & 3.27  &  13.58  & 22.93  & 35.64  & 23.73  \\
    Mask2Former  & R-50    & 21.85  & 6.16  &  16.82  & 31.65  & 41.18  & 28.26  \\
    \textbf{SOIS (Ours)}  & R-50    & \textbf{23.38}  & \textbf{6.59}  & \textbf{17.35}   & \textbf{34.23}  &  \textbf{41.94} & \textbf{29.24}  \\
    \hline
    Mask2Former  & Swin-B  & 33.27  & 9.34  &  25.21  & 47.80  & 50.81  & 37.49  \\
    \textbf{SOIS (Ours)}  & Swin-B  & \textbf{38.02}  &  \textbf{ 12.31 }    &     \textbf{ 28.64 }  &  \textbf{53.22}      &  \textbf{54.74} &    \textbf{41.78}    \\
    \bottomrule
    \end{tabular}
    }
\vspace{-0.2cm}
\end{table*}

% \begin{table}[]
%     \centering
% \caption{Results of Mask2Former and proposed SOIS on COCO2017-val (none-VOC) and COCO2017-val. The models have been trained on COCO2017-train (VOC). Even though the small number of the classes of training data somewhat limits the model's ability to learn generic instance representations, the proposed SOIS still outperforms the baseline on AR$_{100}$(\%). }
% \resizebox{13cm}{!}{
% \begin{tabular}{c|cccc|cc}
% \hline
% {  Test Dataset} & \multicolumn{4}{c|}{  COCO (NoneVOC)}                                                                                                                      & \multicolumn{2}{c}{ COCO}                                 \\ \hline
% {  Metrics}      & \multicolumn{1}{c|}{ AR$_{100}$(\%)} & \multicolumn{1}{c|}{ AR$_s$(\%)} & \multicolumn{1}{l|}{ AR$_m$(\%)} & \multicolumn{1}{l|}{  AR$_l$(\%)} & \multicolumn{1}{l}{  AR$_{100}$(\%)}   \\ \hline
% {  Mask2Former}  & \multicolumn{1}{l}{  9.21}   & \multicolumn{1}{l|}{  4.56}   & \multicolumn{1}{l|}{  8.79}   & \multicolumn{1}{l|}{  19.30}  & \multicolumn{1}{l}{ 37.22}  \\ \hline
% {  SOIS}         & \multicolumn{1}{l}{  11.03}  & \multicolumn{1}{l|}{  4.87}   & \multicolumn{1}{l|}{  9.24}   & \multicolumn{1}{l|}{  26.81}  & \multicolumn{1}{l}{ 38.34}  \\ \hline
% \end{tabular}
% }
% \label{tab: voc-to-novoc}
% \end{table}

\paragraph{Evaluation metrics}The Mean Average Recall (AR) and Mean Average Precision (AP)~\cite{coco} are utilized to measure the performance of approaches in a class-agnostic way.

\subsection{Fully-supervised experimental setting}

\begin{wraptable}{r}{7cm}
\renewcommand\arraystretch{1.0}
    \centering
\caption{Results of \textbf{COCO2017-train(VOC) $\rightarrow$ COCO2017-val(none-VOC)} intra-set evaluation.}
\vspace{-0.2cm}
%Even though the small number of the classes of training data somewhat limits the model's ability to learn generic instance representations, the proposed SOIS still outperforms the baseline on AR$_{100}$(\%). }
% \vspace{-0.3cm}
\resizebox{6cm}{!}{
\begin{tabular}{l|cccc|cc}
\hline
% {  Test Dataset} & \multicolumn{4}{c}{  COCO (NoneVOC)}                                                                                                                                                     \\ \hline
{  Metrics}      & \multicolumn{1}{c|}{ AR$_{100}$} & \multicolumn{1}{c|}{ AR$_s$} & \multicolumn{1}{l|}{ AR$_m$} & \multicolumn{1}{l}{  AR$_l$}   \\ \hline
{  Mask2Former}  & \multicolumn{1}{l|}{  9.21}   & \multicolumn{1}{l|}{  4.56}   & \multicolumn{1}{l|}{  8.79}   & \multicolumn{1}{l}{  19.30}    \\ \hline
{  SOIS}         & \multicolumn{1}{l|}{  11.03}  & \multicolumn{1}{l|}{  4.87}   & \multicolumn{1}{l|}{  9.24}   & \multicolumn{1}{l}{  26.81}   \\ \hline
\end{tabular}
}
\label{tab: voc-to-novoc}
\end{wraptable}
\paragraph{Intra-dataset evaluation} UVO is the largest open-world instance segmentation dataset. Its training and test images are selected from the same domain, while they do not have any overlap. Here, we perform the leaning process of SOIS on the UVO-train subset and conduct the test experiments on the UVO-val subset.  Besides, we split the COCO dataset into 20 seen (VOC) classes and 60 unseen (none-VOC) classes. We train a model only on the annotation of 20 VOC classes and test it on the 60 none-VOC class, evaluating its ability of discovering novel objects.

\paragraph{Cross-dataset evaluation} Open-world setting assumes that the instance can be novel classes in the target domain. Therefore, it is essential for the OWIS method to handle the potential domain gap with excellent generalization ability. Cross-dataset evaluation, in which training and test data come from different domains, is necessary to be conducted. Here, we first train the proposed SOIS model and compared methods on the COCO-train subset, while testing them on the UVO-val dataset to evaluate their generalizability. Then we extend the experiments to an autonomous driving scenario, training the models on the Cityscapes~\cite{cityscapes} dataset and evaluating them on the Mapillary~\cite{mapillary}.Cityscape have 8 foreground classes , while Mapillary contains 35 foreground classes including vehicles, animals, trash can, mailbox, etc.

\begin{table*}[t]
\renewcommand\arraystretch{1.0}
\setlength\tabcolsep{4pt}
    \centering
    \vspace{-0.2cm}
    \caption{Results of \textbf{COCO2017-train $\rightarrow$ UVO-val} cross-dataset evaluation.}  
    \vspace{-0.2cm}
    \label{table2}
    \resizebox{11cm}{!}{
    \begin{tabular}{l|l|ccccccc}
    \toprule
    % \multirow{2}*{Metric} & AR$_{100}$  & AP$_{100}$ & AP$_s$ & AP$_m$ & AP$_l$\\
    %       & (\%) & (\%) & (\%) & (\%) & (\%) \\
    Metric & Backbone&AR$_{100}$ & AP$_{100}$ & AP$_s$ & AP$_m$ & AP$_l$ \\
          
    %\midrule
    \hline
    MaskRCNN    &R-50      & 38.17  & 19.05 &  6.27  &  13.15  & 28.05  \\
    LDET&   R-50              & 42.63  & 21.27 &  5.66  &  17.52&  18.38 \\
    GGN &R-50           & 43.30  & 20.30    & \textbf{8.70}    & 18.20     & 27.30  \\\hline
    % SOLO V2-R50 & 39.41 & 22.25 & 5.56 & 14.18 & 34.12 \\ 
    % SOLO V2-R50+SOIS & 42.52 & 25.04 & 6.77 & 16.90 & 38.33 \\ \hline
    Mask2Former&R50    & 48.71       &  25.24     &   6.46     &   16.09      &   40.37   \\
    \textbf{SOIS (Ours)}    & R-50 & \textbf{51.28}   & \textbf{27.62} & 7.80   &  \textbf{18.61}  & \textbf{43.42}  \\
    \hline
    Mask2Former &Swin-B& 51.38  & 28.16 &     7.29   &  18.91       &   45.48   \\
    \textbf{SOIS(Ours)} &Swin-B  & \textbf{54.86}   & \textbf{32.21} & \textbf{9.03}   &  \textbf{21.92}  & \textbf{50.69}\\
    \bottomrule
    \end{tabular}
    }
\vspace{-0.2cm}
\end{table*}

\subsection{Fully-supervised experimental results}
\paragraph{Intra-dataset evaluation}
The results are illustrated in Table~\ref{table1}. The single-stage approaches based on the mask classification framework perform better than other two-stage methods. Among them, our proposed SOIS achieves a significant performance improvement over the Mask2Former baseline, which is 4.75\% in $AP_{100}$ and 3.93\% in $AR_{100}$ when using the Swin-B backbone. For VOC$\rightarrow$none-VOC setting, the experimental results are shown in Table~\ref{tab: voc-to-novoc}, which verified that our proposed method can improve the performance for all instances, especially large ones.

\begin{wraptable}{r}{7cm}
%\begin{table*}[h]
% \renewcommand\arraystretch{1.1}
% \setlength\tabcolsep{8pt}
    \centering
    \caption{Results of SOIS with SOLOV2 structure ( \textbf{UVO-train$\rightarrow$ UVO-val}). }
    \label{tablesolo}
    \resizebox{7cm}{!}{
    \begin{tabular}{l|l|ccccc}
    \toprule
    % Test dataset & Backbone&\multicolumn{5}{c}{UVO} \\
    %\midrule
    % \multirow{2}*{Metric} & AR$_{100}$  & AP$_{100}$ & AP$_s$ & AP$_m$ & AP$_l$\\
    %       & (\%) & (\%) & (\%) & (\%) & (\%) \\
    Metric & Backbone &AR$_{100}$ & AP$_{100}$ & AP$_s$ & AP$_m$  & AP$_l$\\          
    %\midrule
    \hline
    SOLO V2 &R-50 & 39.41 & 22.25 & 5.56 & 14.18 & 34.12 \\ 
    SOLO V2SOIS &R-50& 42.52 & 25.04 & 6.77 & 16.90 & 38.33 \\ 

    \bottomrule
    \end{tabular}
     }
%\end{table*}
\end{wraptable}
\paragraph{Cross-dataset evaluation} 
For the COCO$\rightarrow$UVO task, according to Table~\ref{table2}, it is clear that the proposed SOIS outperforms all previous methods, achieving a new state-of-the-art $AR_{100}$ at 54.86\% which is 11.56\% higher than previous state-of-the-art method GGN~\cite{ggn}. We also applied the proposed techniques to another classic one-stage method SOLO V2~\cite{solov2}. The experimental results in Table~\ref{tablesolo} show that it improves $AR_{100}$ and $AP_{100}$ by 3.11\% and 2.79\% compared to SOLO V2. For the  Cityscape$\rightarrow$ Mapillary task, the overall $AP$ and $AR$ of SOIS still surpass the performance of other state-of-the-art methods(in Table~\ref{tab: ctscp}), which demonstrates the effectiveness of our proposed techniques. We show some of the COCO$\rightarrow$UVO visualization results in Figure~\ref{fig:fig3} to qualitatively demonstrate the superiority of our method. Please refer to the supplementary material for more qualitative examples.

% \subsubsection{Mask2Former itself is not enough.} \label{sec:not-just-mask2form} Note that single-stage Mask2Former does not always beat MaskRCNN-based methods. Table~\ref{tab: exp_auto} shows a case where the MaskRCNN-based method LDET outperforms Mask2Former on AP. After introducing our proposed module, Mask2Former beats the MaskRCNN-based method with a clear advantage. This result has verified that our proposed method achieves the superior performance, which is not just because we chose the better backbone; the more important reason behind this is that our proposed framework does work efficiently. 

\begin{figure*}[t]
	\centering
% 	\includegraphics[width=0.17\linewidth]{figs/fkVMf-qLEI_210.png}
% 	\includegraphics[width=0.17\linewidth]{figs/maskrcnn--fkVMf-qLEI_210.png}
% 	\includegraphics[width=0.17\linewidth]{figs/ldet--fkVMf-qLEI_210.png}
% 	\includegraphics[width=0.17\linewidth]{figs/mask2former--fkVMf-qLEI_210.png}
% 	\includegraphics[width=0.17\linewidth]{figs/highlightfig/1.png}
% 	%\includegraphics[width=0.245\linewidth]{figs/grow--fkVMf-qLEI_210.png}
% 	\quad
%     \includegraphics[width=0.17\linewidth]{figs/Fs-zEnMYgc_120.png}
% 	\includegraphics[width=0.17\linewidth]{figs/maskrcnn--Fs-zEnMYgc_120.png}
% 	\includegraphics[width=0.17\linewidth]{figs/ldet--Fs-zEnMYgc_120.png}
% 	\includegraphics[width=0.17\linewidth]{figs/mask2former--Fs-zEnMYgc_120.png}
% 	\includegraphics[width=0.17\linewidth]{figs/highlightfig/2.png}
% 	%\includegraphics[width=0.245\linewidth]{figs/grow--fkVMf-qLEI_210.png}
% 	\quad
% 	\includegraphics[width=0.17\linewidth]{figs/1brKJdL-iM_180.png}
% 	\includegraphics[width=0.17\linewidth]{figs/maskrcnn-1brKJdL-iM_180.png}
% 	\includegraphics[width=0.17\linewidth]{figs/ldet-1brKJdL-iM_180.png}
% 	\includegraphics[width=0.17\linewidth]{figs/mask2former-1brKJdL-iM_180.png}
% 	\includegraphics[width=0.17\linewidth]{figs/highlightfig/3.png}
% 	%\includegraphics[width=0.245\linewidth]{figs/grow--fkVMf-qLEI_210.png}
% 	\quad
	\includegraphics[width=0.17\linewidth]{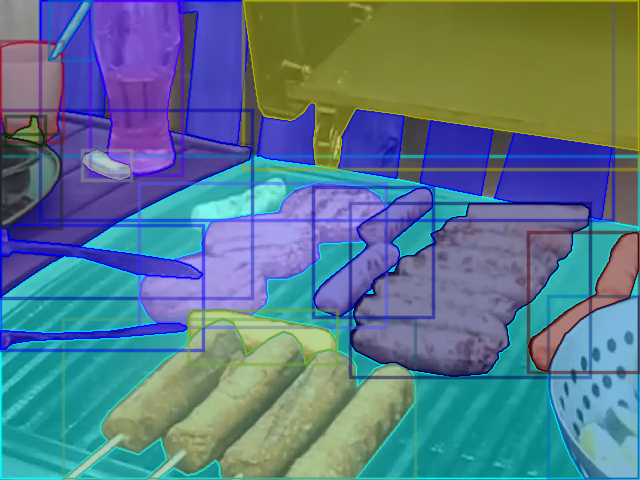}
	\includegraphics[width=0.17\linewidth]{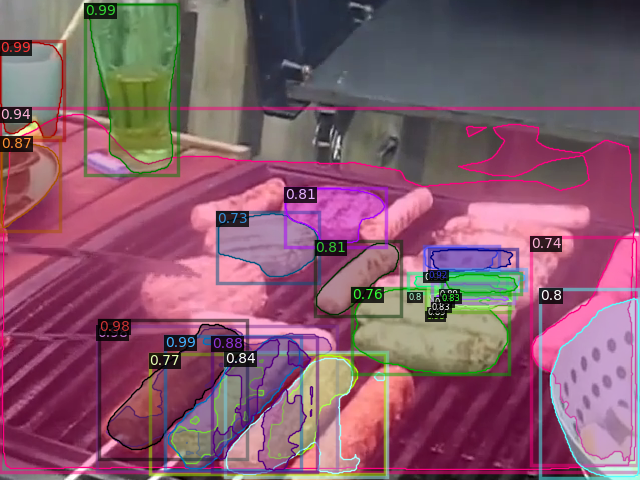}
	\includegraphics[width=0.17\linewidth]{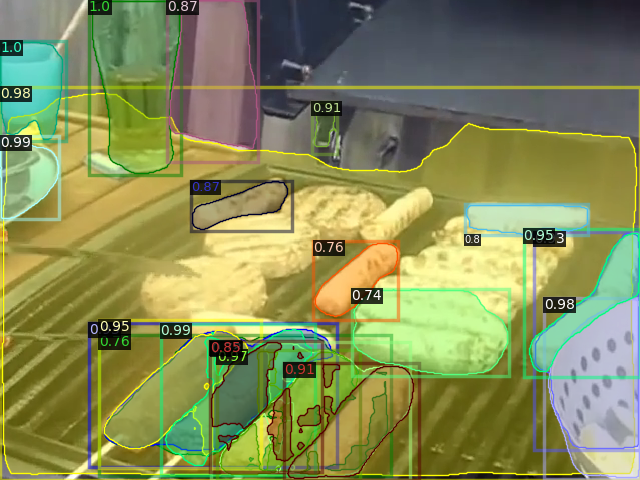}
	\includegraphics[width=0.17\linewidth]{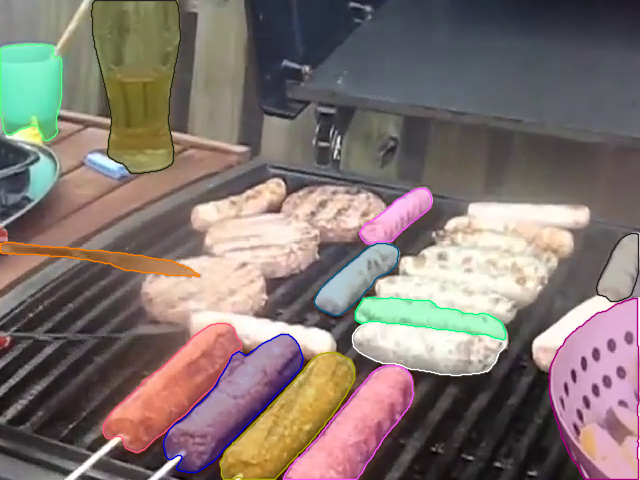}
	\includegraphics[width=0.17\linewidth]{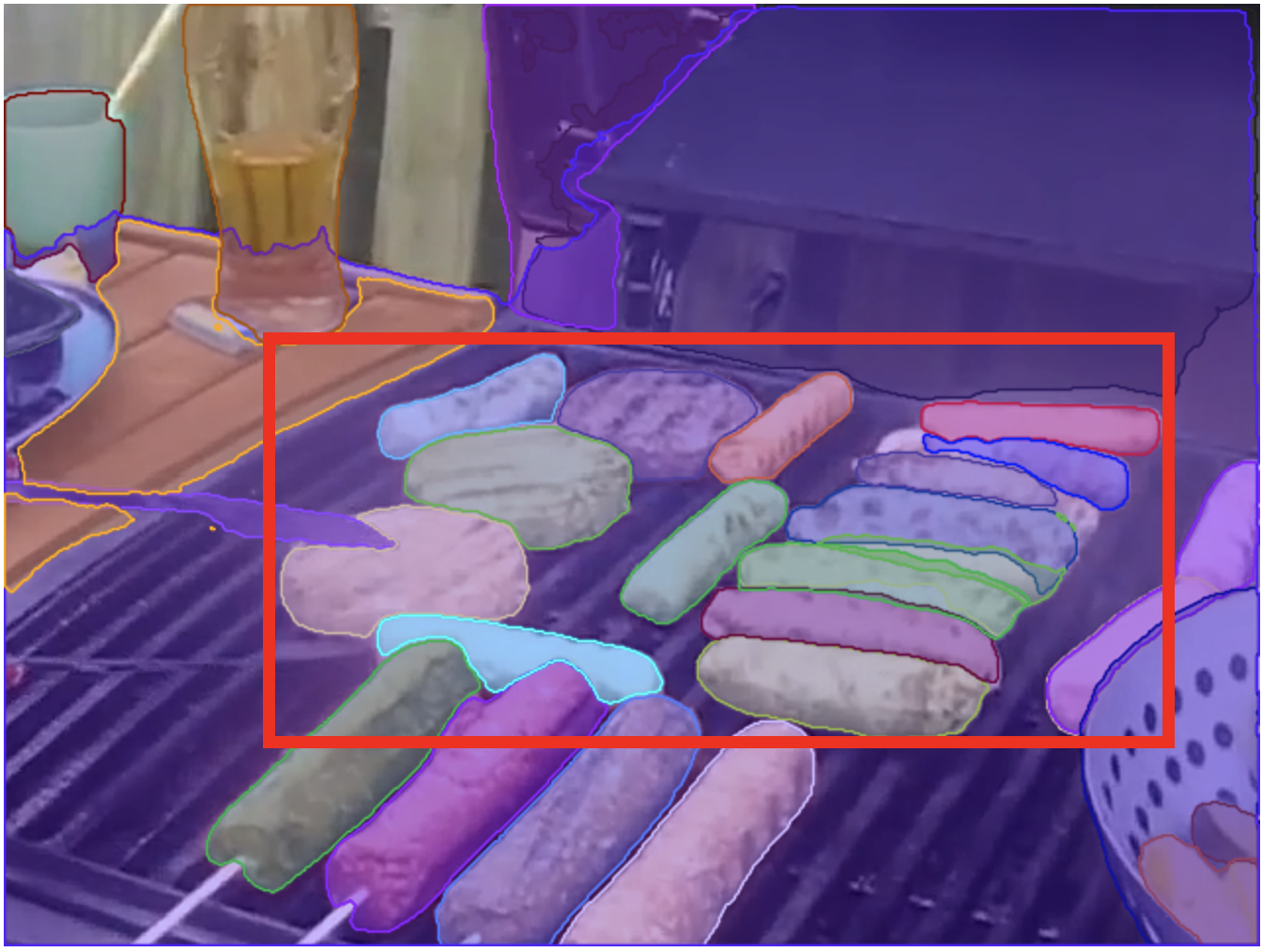}
	\quad
	\subfloat[Groundtruth]{\includegraphics[width=0.17\linewidth]{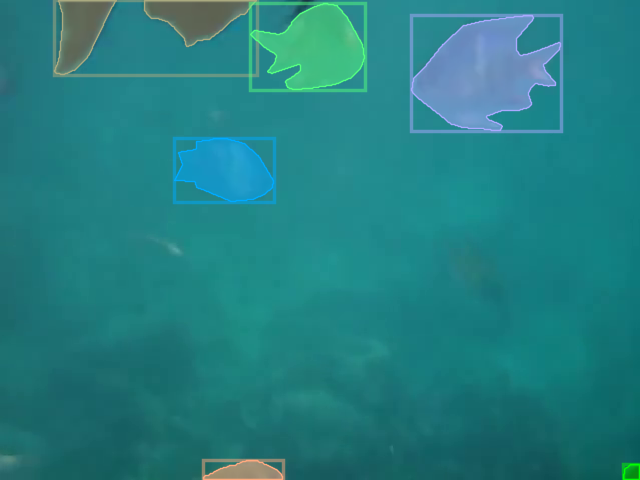}}
	\subfloat[MaskRCNN]{\includegraphics[width=0.17\linewidth]{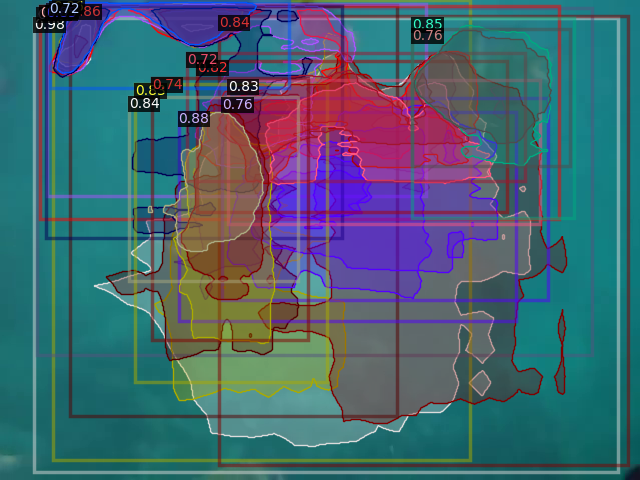}}
	\subfloat[LDET]{\includegraphics[width=0.17\linewidth]{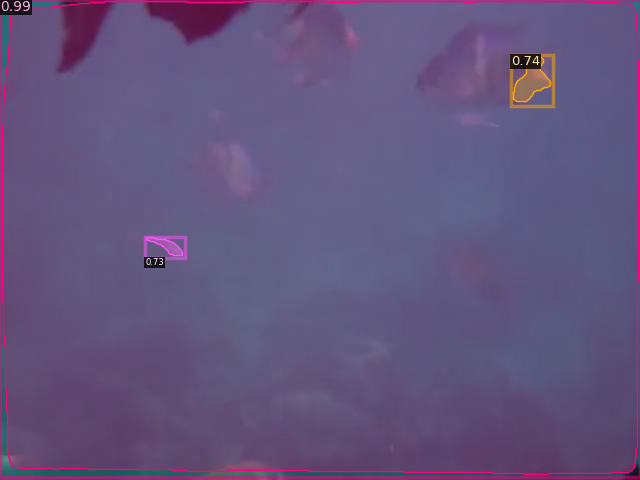}}
	\subfloat[Mask2Former]{\includegraphics[width=0.17\linewidth]{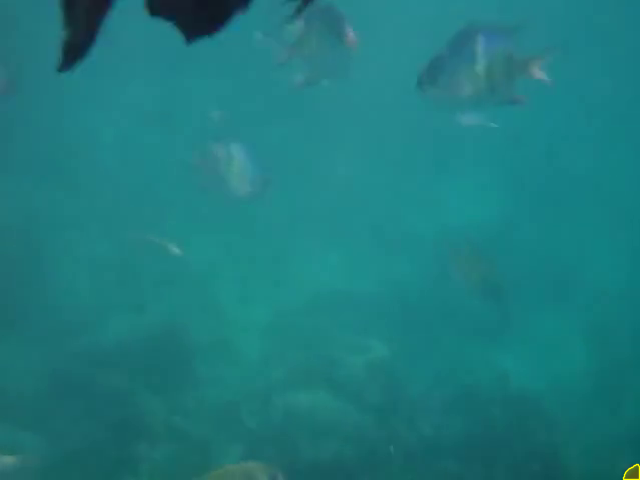}}
	\subfloat[SOIS (Ours)]{\includegraphics[width=0.17\linewidth]{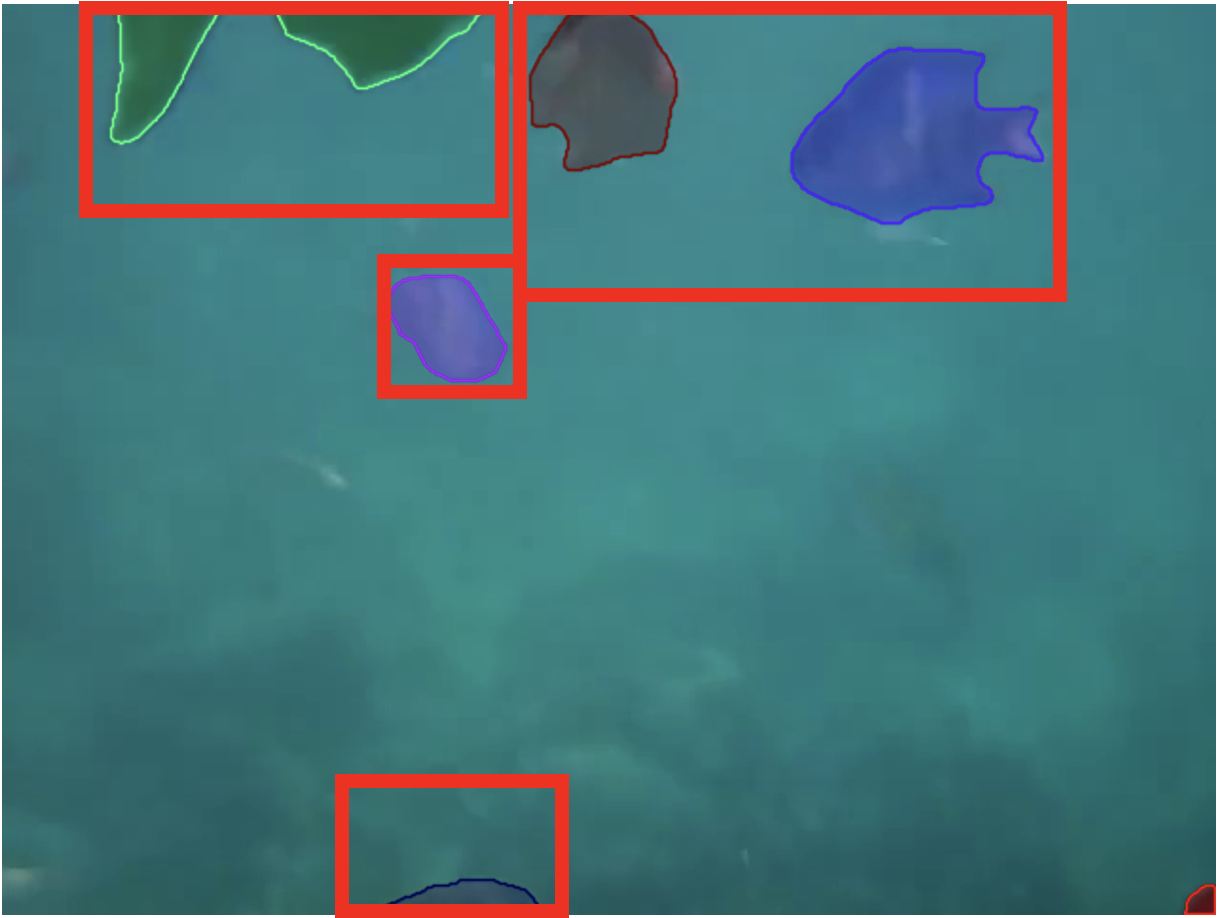}}
	%\subfloat[SOIS (Ours)]{\includegraphics[width=0.245\linewidth]{figs/grow--jJvz1Ul23o_90.png}}   
	\caption{Visualization results of COCO$\rightarrow$UVO cross-dataset evaluation. The predicted boxes of two-stage methods MaskRCNN and LDET are also drawn. Proposed SOIS can discover both unlabeled object (first row) and unseen class of instances (second row) as shown in \textcolor{red}{red boxes}.}
\label{fig:fig3}
	
\end{figure*}

\subsection{Ablation study}

We perform cross-dataset and intra-dataset ablation studies to analyze the effectiveness of each component in the proposed SOIS, including the  foreground prediction branch and the cross-task consistency loss. We also try combinations of the pseudo-label generation strategy and our cross-task consistency loss to investigate the individual and synergetic effects of them. Using the SwinB backbone, these models are trained on the COCO-train subset and the UVO-train subset, respectively. The metrics reported in Table~\ref{tab:ablation_study} are tested on the UVO-val dataset. 

\begin{table}
%\begin{table*}[]
\renewcommand\arraystretch{1.0}
\setlength\tabcolsep{4.0pt}
    \centering
    \caption{Ablation results of the proposed components by cross-dataset and intra-dataset evaluations. Foreground prediction (FP), Cross-task consistency (CTC) loss, Pseudo label (PL).}
    
    \label{sample-table}
    \resizebox{11cm}{!}{
    \begin{tabular}{c|c|c|cc|cc}
    \toprule
    \multicolumn{3}{c|}{Component} &\multicolumn{2}{c|}{Train on COCO} & \multicolumn{2}{c}{Train on UVO} \\
    \hline
    FP & CTC loss  & PL  & AP$_{100}$(\%) & AR$_{100}$(\%) & AP$_{100}$(\%) & AR$_{100}$(\%)\\
    \hline
               &           &           &  28.65  &  51.54  & 35.12 & 51.39   \\
    \checkmark &           &           &  29.02  &  51.60 & 35.55 &   51.73 \\
               &           &\checkmark &  30.09  &  52.97 & 32.94 &   51.64 \\
    \checkmark &\checkmark &           &  31.39  &  53.83 & \textbf{38.02} &   \textbf{54.74} \\
    \checkmark &           &\checkmark &  30.17  &  52.98 & 33.35 &   50.90 \\
    \checkmark &\checkmark &\checkmark &  \textbf{32.21}  & \textbf{54.86} &37.71 &52.27 \\
    \bottomrule
    \end{tabular}
    }

    \label{tab:ablation_study}
%\end{table*}
\end{table}
\paragraph{Effectiveness foreground prediction branch} Table~\ref{tab:ablation_study} shows that although a separate foreground prediction branch can guide the method to optimize towards the direction of discovering foreground pixels, it only slightly boosts the performance.
% \begin{figure}[]
% 	\begin{center}
% 		\includegraphics[width=0.95\linewidth]{figs/Figure-semi.png}
% 	\end{center}
% 	\caption{ Overall framework of the proposed SOIS. The mask prediction branch generates the predicted masks, while the objectness prediction branch computes the objectness score for each mask. The auxiliary foreground prediction branch segments a foreground region to guide the optimization of other two branches.}
% 	\label{fig:figsemi}
% \end{figure}

\paragraph{Effectiveness of cross-task consistency loss} Cross-task consistency loss has a positive effect on both sparse annotated (COCO) and dense annotated (UVO) training dataset. The values of $AP_{100}$ and $AR_{100}$ increase significantly ( 2.74\% $\uparrow$ and 2.49\% $\uparrow$ on COCO while 2.90\% $\uparrow$ and 3.35\%$\uparrow$ on UVO) after applying the cross-task consistency loss as well as the foreground prediction branches together. This result outperforms the SOIS counterpart with only pseudo-labeling, showing our effectiveness. In addition, jointly utilizing our cross-task consistency loss as well as the pseudo-labeling strategy leads to performance improvements on two settings, which demonstrates the synergistic effect of both approaches.

\paragraph{Effectiveness of pseudo-labeling} Pseudo-labeling is not always necessary and powerful for any types of datasets. As shown in Table~\ref{tab:ablation_study}, the $AP_{100}$ and $AR_{100}$ of the COCO trained model increase by 1.35\% and 0.78\%, respectively, after applying the pseudo-label generation. However, pseudo-labeling causes a performance degradation (e.g. 2.18\%$\downarrow$ in $AP_{100}$) to a model trained in the UVO dataset. Compared with COCO, the UVO dataset is annotated more densely. We conjecture that the background annotations of UVO are more reliable than those of COCO, where carefully selected pseudo-labels are more likely to represent unlabeled objects. The generated pseudo-labels of UVO contain higher noises than those of COCO. These additional noisy labels mislead the model training.

\subsection{Semi-supervised learning experiment}

\begin{figure}[t]
    \begin{minipage}[t]{0.5\textwidth}
    \setlength{\belowcaptionskip}{1.7pt}
    \begin{center}
    \makeatletter\def\@captype{table}\makeatother\caption{\small{Results of our SOIS and classic semi-supervised method on UVO-val.}}
    \setlength{\tabcolsep}{2.8mm}{
    \resizebox{7cm}{!}{
     \begin{tabular}{l|cccc}
        \toprule
        Training Data   & \multicolumn{4}{c}{UVO-train with 30\% annotation}  \\
        \hline
         Method & Fully-SOIS$_{30}$   &Mean Teacher  &Pseudo Labeling   & Semi-SOIS$_{30}$      \\
         \hline
         AP$_{100}$(\%)  & 21.67  &21.95 &22.77 & 25.03   \\
         AR$_{100}$(\%)  & 40.09  &40.82  &41.56  & 45.42   \\
         \bottomrule
       \end{tabular}}
    \label{table6}}
    \end{center}
    \end{minipage}
    \quad
    \begin{minipage}[t]{0.47\textwidth}
    \setlength{\belowcaptionskip}{1.7pt}
    \begin{center}
    \makeatletter\def\@captype{table}\makeatother\caption{\small{Results of our SOIS and recent end to end method on UVO-val.}}
    \setlength{\tabcolsep}{4.2mm}{
    \resizebox{7cm}{!}{
    \begin{tabular}{l|cccc}
        \toprule
        Training Data   & \multicolumn{4}{c}{UVO-train with 50\% annotation} \\
        \hline
         Method  & LDET$_{50}$ &Mask2Former$_{50}$ &Fully-SOIS$_{50}$       & Semi-SOIS$_{50} $      \\
         \hline
         AP$_{100}$(\%)    &10.61 &19.49 & 22.86    & 25.22  \\
         AR$_{100}$(\%)  &25.08 &38.08  & 41.44    & 47.56  \\
         \bottomrule
       \end{tabular}}
    \label{table7}}
    \end{center}
    \end{minipage}
\end{figure}

\paragraph{Experimental setting} We have divided the UVO-train dataset into the labeled subset $D_l$ and the unlabeled subset $D_u$. Semi-supervised model Semi-SOIS is optimized as described in Section 3.6 on $D_l\cup D_U$, while the fully-supervised method Fully-SOIS is trained merely on the $D_L$. To ensure the comprehensiveness of the experiment, two different data division settings are included in our experiments:\{$D_L$=30\%, $D_u$=70\%\} and \{$D_L$=50\%, $D_u$=50\%\}. The backbone applied here is Swin-B. We also implemented the classic Mean teacher model and a simple pseudo-label method based on the Mask2Former to perform comparison.

\begin{wrapfigure}{r}{5.5cm}
\includegraphics[width=5cm, height=5.0cm]{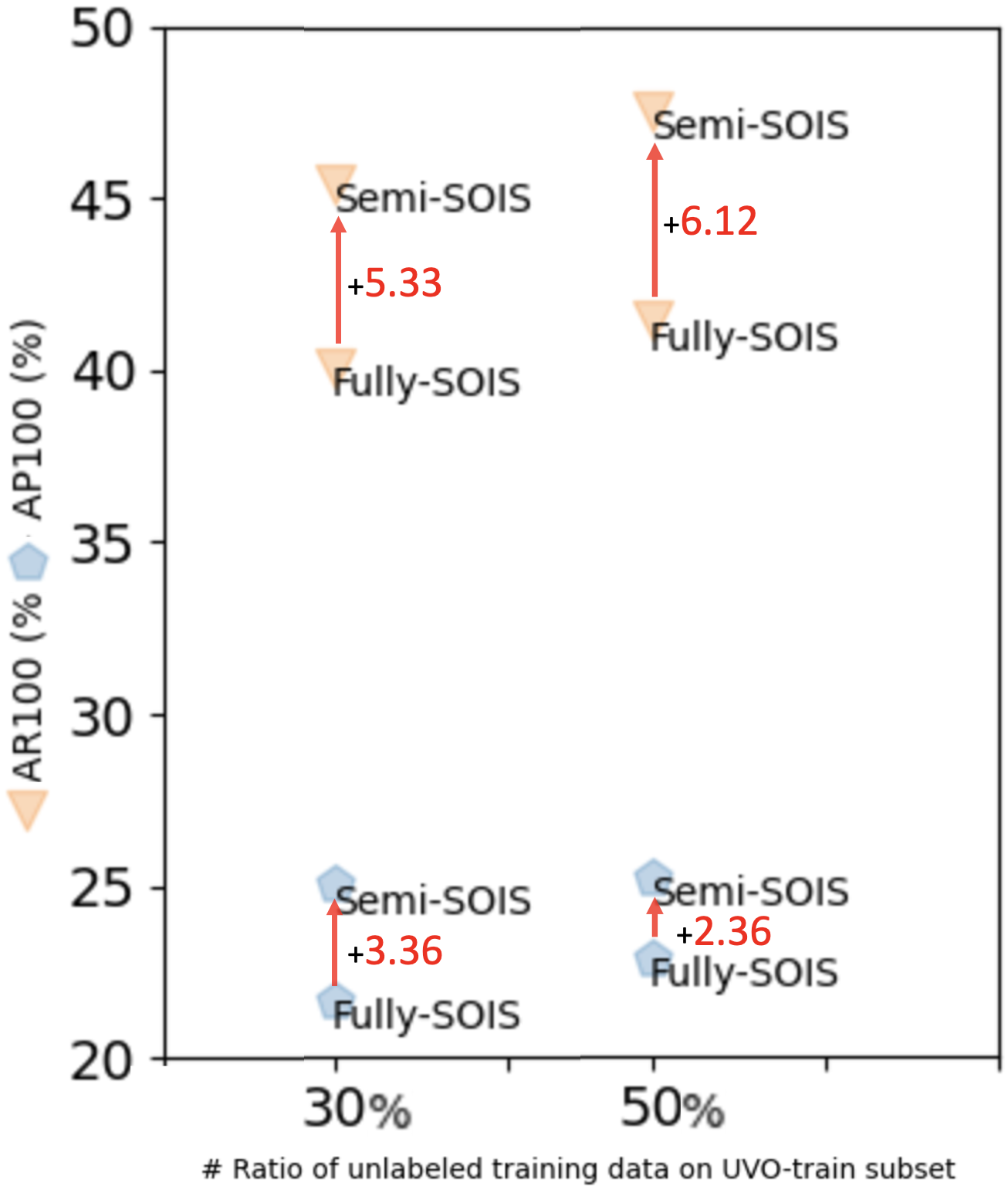}
\caption{Comparison between fully-SOIS and semi-SOIS}
%\caption{ Overall framework of the proposed SOIS. The mask prediction branch generates the predicted masks, while the objectness prediction branch computes the objectness score for each mask. The auxiliary foreground prediction branch segments a foreground region to guide the optimization of other two branches.}
\label{fig:figsemi}
\end{wrapfigure}
\paragraph{Results and analysis} As presented in Figure~\ref{fig:figsemi}, the Semi-$SOIS_{50}$ model trained on the UVO with 50\% annotated data outperforms the Semi-$SOIS_{30}$ model leaning with 30\% labeled training images. However, the performance increase between the Semi-$SOIS_{30}$ and Semi-$SOIS_{50}$ is slight. In addition, Semi-$SOIS_{30}$ improves Fully-$SOIS_{30}$ by 3.36\% and 5.33\% in  $AP_{100}$ and $AR_{100}$, respectively. Compared to Fully-$SOIS_{50}$, Semi-$SOIS_{50}$ still achieves significant advantages (2.36\% in $AP_{100}$ and 6.12\% in $AR_{100}$). These results reflect that cross-task consistency loss has the ability to extract information from unlabeled data and facilitates model optimization in the semi-supervised setting. It is notable that the results of Semi-$SOIS_{30}$ are even better than those of Fully-$SOIS_{50}$. This illustrates that the information dug out by the cross-task consistency loss from the remaining 70\% unlabeled data is more abundant than that included in 20\% fully-labeled data. Therefore, our algorithm can achieve better performance with fewer annotations. This characteristic is promising in solving the OWIS problem. In addition, we also compared the semi-$SOIS$ with classic semi-supervised method and recent end to end segmentation method. The results in Table~\ref{table6} and ~\ref{table7} show our advantages over the compared methods.

\begin{wraptable}{r}{7cm}
% \begin{table}[!ht]
% \renewcommand\arraystretch{1.1}
% \setlength\tabcolsep{13pt}
    \centering
    \caption{Cross-set evaluation on autonomous driving scenes. Results of \textbf{Cityscapes $\rightarrow$ Mapillary}.}
    \label{tab: ctscp}
    \resizebox{7cm}{!}{
    \begin{tabular}{l|ccccccc}
    \hline
        Method & MaskRCNN & LDET & Mask2Former & OSIS(Ours) \\ \hline
        AP(\%) & 7.3 & 7.8 & 7.6 & 8.4  \\ \hline
        AR$_{10}$(\%) & 6.1 & 5.5 & 7.0 & 7.5  \\ \hline
    \end{tabular}
    }
%\end{table}
\end{wraptable}
% \begin{table}[]
% \renewcommand\arraystretch{1.1}
% \setlength\tabcolsep{12pt}
%     \centering
%     \caption{\small Results of models trained with only labeled data (LDET, Mask2Former, Fully-SOIS) and trained with both labeled and unlabeled data(Mean Teacher, Pseudo Labeling, Semi-SOIS) on UVO-val dataset.}
%     \label{table4}
%     %\vspace{0.2cm}
%     \resizebox{14cm}{!}{
%       \begin{tabular}{l|cccc|cccc}
%         \toprule
%         Training Data   & \multicolumn{4}{c|}{30\% labeled data   in UVO} & \multicolumn{4}{c}{50\% labeled  data  in UVO} \\
%         \hline
%          Method & Fully-SOIS$_{30}$   &Mean Teacher  &Pseudo Labeling   & Semi-SOIS$_{30}$       & LDET$_{50}$ &Mask2Former$_{50}$ &Fully-SOIS$_{50}$       & Semi-SOIS$_{50} $      \\
%          \hline
%          AP$_{100}$(\%)  & 21.67  &21.95 &22.77 & 25.03  &10.61 &19.49 & 22.86    & 25.22  \\
%          AR$_{100}$(\%)  & 40.09  &40.82  &41.56  & 45.42 &25.08 &38.08  & 41.44    & 47.56  \\
%          \bottomrule
%       \end{tabular}

%      }
% \end{table}

\section{Conclusion}
%\paragraph{Results and analysis}

 This paper proposes the first single-stage framework (SOIS) for the open-world instance segmentation task. Apart from predicting the instance mask and objectness score, our framework introduces a foreground prediction branch to segment the regions belonging to any instance. Utilizing the outputs of this branch, we propose a novel cross-task consistency loss to enforce the foreground prediction to be consistent with the prediction of the instance masks. We experimentally demonstrate that this mechanism alleviates the problem of incomplete annotation, which is a critical issue for open-world segmentation. Our extensive experiments demonstrate that SOIS outperforms state-of-the-art methods by a large margin on typical datasets. We further demonstrate that our cross-task consistency loss can utilize unlabeled images to obtain some performance gains for a semi-supervised instance segmentation. This is an important step toward reducing laborious and expensive human annotation. 

\begin{ack}
Mike Zheng Shou is supported only by the National Research Foundation, Singapore under its NRFF award NRF-NRFF13-2021-0008.
\end{ack}
\newpage
\appendix
\section{Appendix}
In this appendix, we provide the architecture of the foreground prediction branch (in Figure~\ref{fig:fig1}) and detailed experimental settings first. Then some annotations in UVO dataset are visualized in Figure ~\ref{fig:fig3} to show the challenges of open world instance segmentation. Finally, additional visualization results of proposed SOIS are shown in Figure~\ref{fig:fig4}.

\begin{figure*}[h]
	\begin{center}
		\includegraphics[width=1\linewidth]{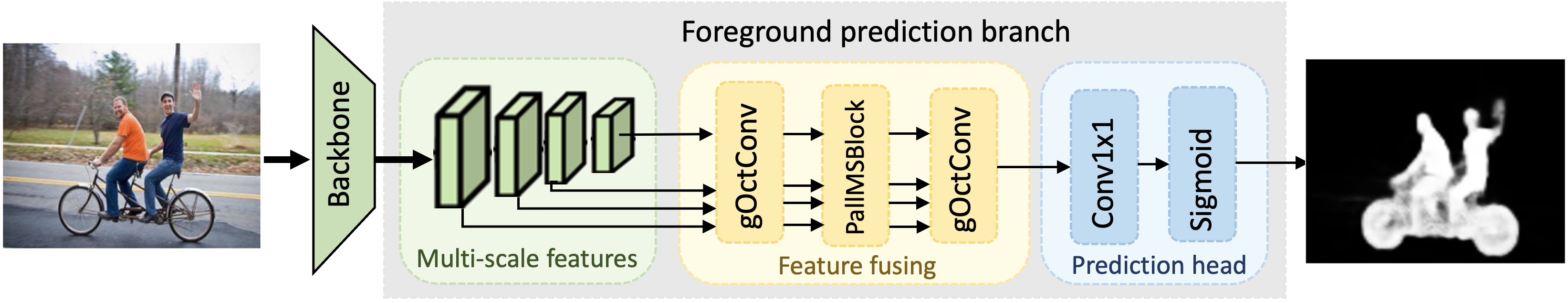}
	\end{center}
	\caption{\textbf{Architecture of foreground prediction branch}. Multi-scale features extracted from backbone are fed into the feature fusing module to exchange and fuse the multi-scale information. Then a fused feature is sent to the prediction head to predict the final foreground map. Considering the efficiency, we follow ~\cite{gao2020highly} to introduce the gOctConv ~\cite{gao2020highly} and PallMSBlock~\cite{gao2020highly}  to perform feature fusing.}	
	\label{fig:fig1}
\end{figure*}
\subsection{Detailed experimental settings}

\paragraph{Implementation details}
For feature extracting, we obtain the multi-scale features  through a sequential backbone network~\cite{liu2021Swin,he2016deep}, and FPN~\cite{fpn}. The multi-scale features contain D-dimensional feature maps with resolutions of 1/4,  1/8, 1/16, and 1/32. In the pixel decoder module, six MSDeformAttn layers are employed, while the transformer decoder have three layers with 100 queries by default.

In fully-supervised learning, the total loss $L_f$ can be formulated as: $L_f=\alpha L_m+\beta L_p+\gamma L_c +\omega L_o$. We set the weight $\alpha$ of mask loss ($L_m$) to 5.0, the weight $\beta$ of foreground loss ($L_p$) to 1.0, the weight $\gamma$ of cross-task consistency loss ($L_c$) to 1.0 and the weight $\omega$ of objectness loss ($L_o$) to 2.0.

\paragraph{Training settings}
Specifically, AdamW~\cite{adamw} optimizer and the step learning rate schedule are applied to optimize our model. An initial learning rate of 0.0001 and a weight decay of 0.05 are utilized for all backbones. We set a learning rate multiplier of the backbone to 0.1 and we decay the learning rate at 0.9 and 0.95 fractions of the total number of training steps by a factor of 10. For data augmentation, we use the large-scale jittering (LSJ) augmentation with a random scale sampled from range 0.1 to 2.0 followed by a fixed size crop to 1024$\times$1024 on COCO dataset and 640$\times$640 on UVO dataset. Besides, a Cutout~\cite{devries2017improved} strategy that randomly cuts out a region of size [1/8$\cdot$w, 1/8$\cdot$h] to [1/3$\cdot$w, 1/3$\cdot$h] is introduced during training. On COCO dataset, we train our models for $38\times10^4$ iterations with a batch size of 16, while on UVO dataset, we train our models for $12\times10^4$ iterations with the same batch size.

\textbf{SOIS training process with pseudo-labeling on COCO dataset}
\\
\begin{algorithm}[H]  
  \KwData{Image dataset}  
  \KwResult{Proposed SOIS Model $M_u$}  
  initialization the student model $M_u$, and teacher model $M_t$=$M_u$.copy()\; 
\While{Image $i$  $ \notin $ $\varnothing$}{  
      read image $i$ and corresponding groundtruth $gt_i$\;  
      extract backbone feature $X_i$\;
      pred\_masks $\leftarrow$ $M_t$.predictor($X_i$)\;  
      pseudo\_proposals$\leftarrow$ filter\_masks\_with\_confidence(pred\_masks, confidence\_threshold)\;
      pseudo labels $\leftarrow$ filter\_masks\_with\_IOU(pseudo\_proposals, IOU\_threshold)\;
      training labels $\leftarrow$ merge($gt_i$, pseudo labels)\;
      aug\_data$\leftarrow$ Cutout($X_i$, training labels)\; 
      $M_u$ $\leftarrow$ $M_u$.training(aug\_data)\;
      $M_t$ $\leftarrow$ $M_t$.EMA\_update($M_t$,$M_u$)
  }  
  \caption{SOIS training process with pseudo-labeling}  
\end{algorithm} 

\begin{figure*}[]
	\begin{center}
		\includegraphics[width=1\linewidth]{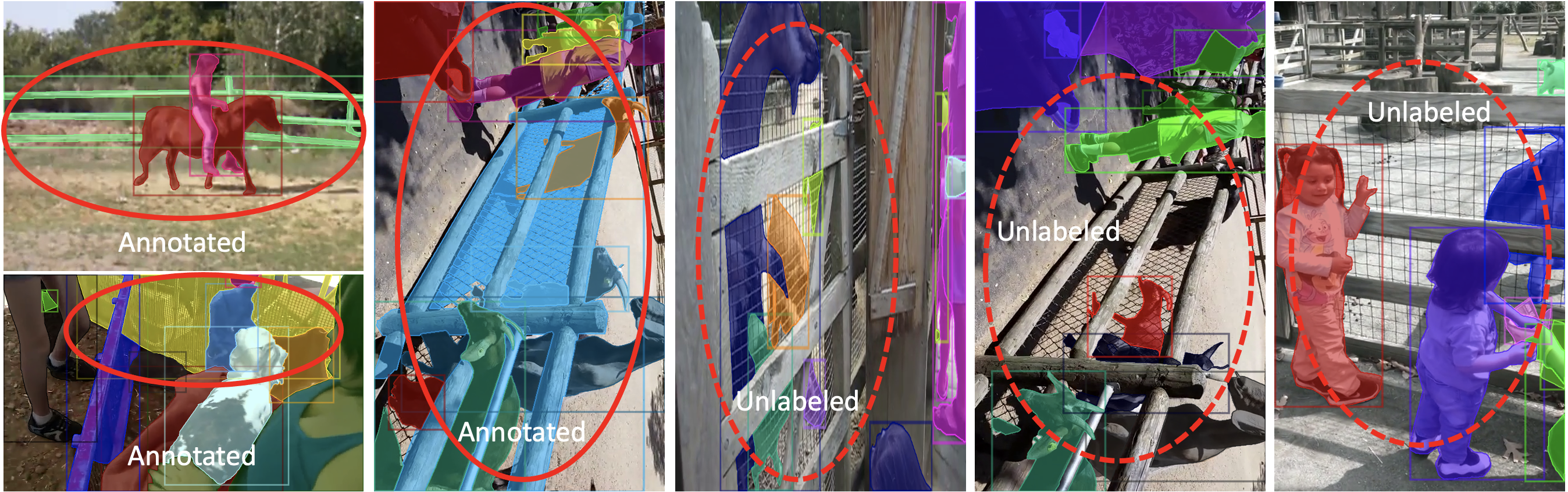}
		\includegraphics[width=1\linewidth]{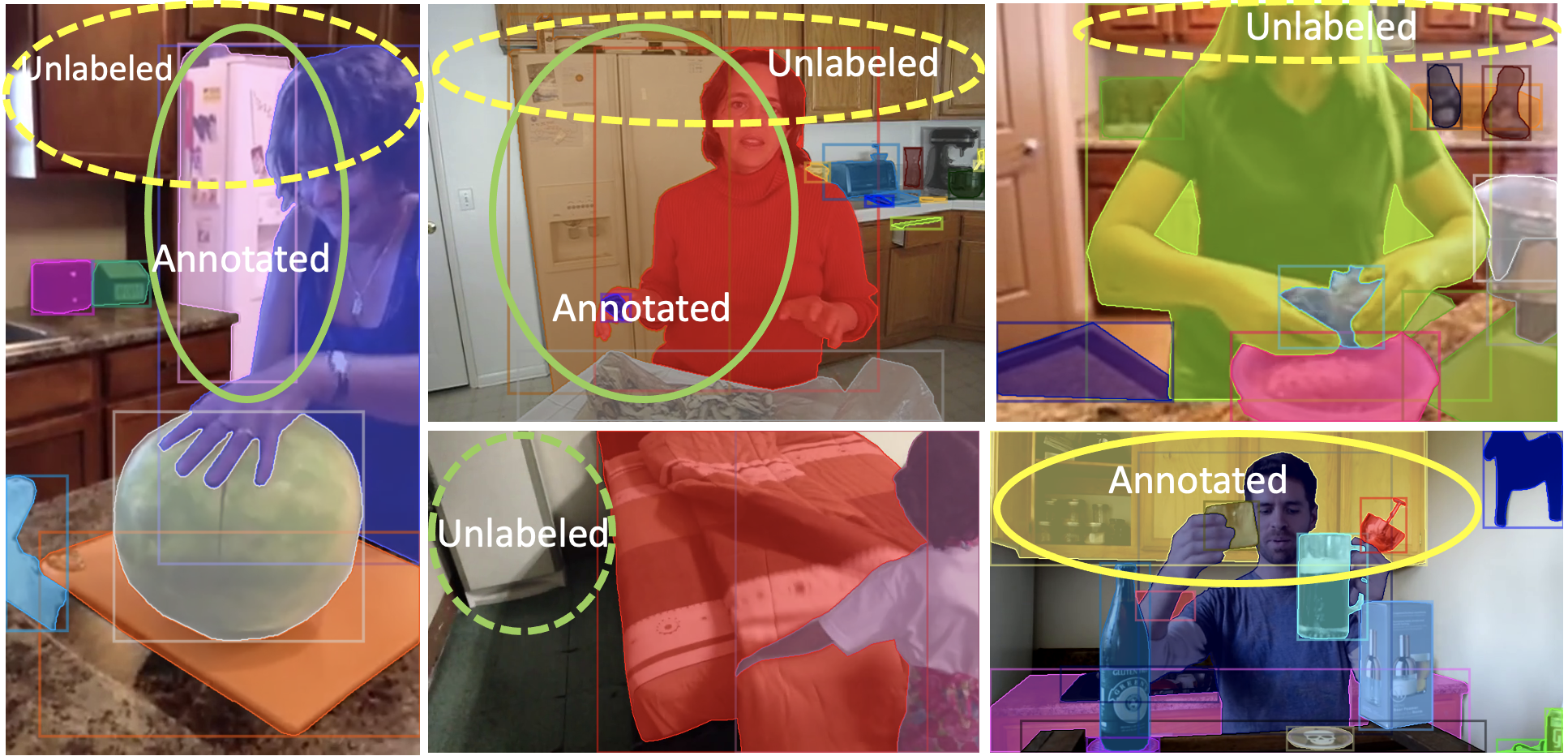}
		\includegraphics[width=1\linewidth]{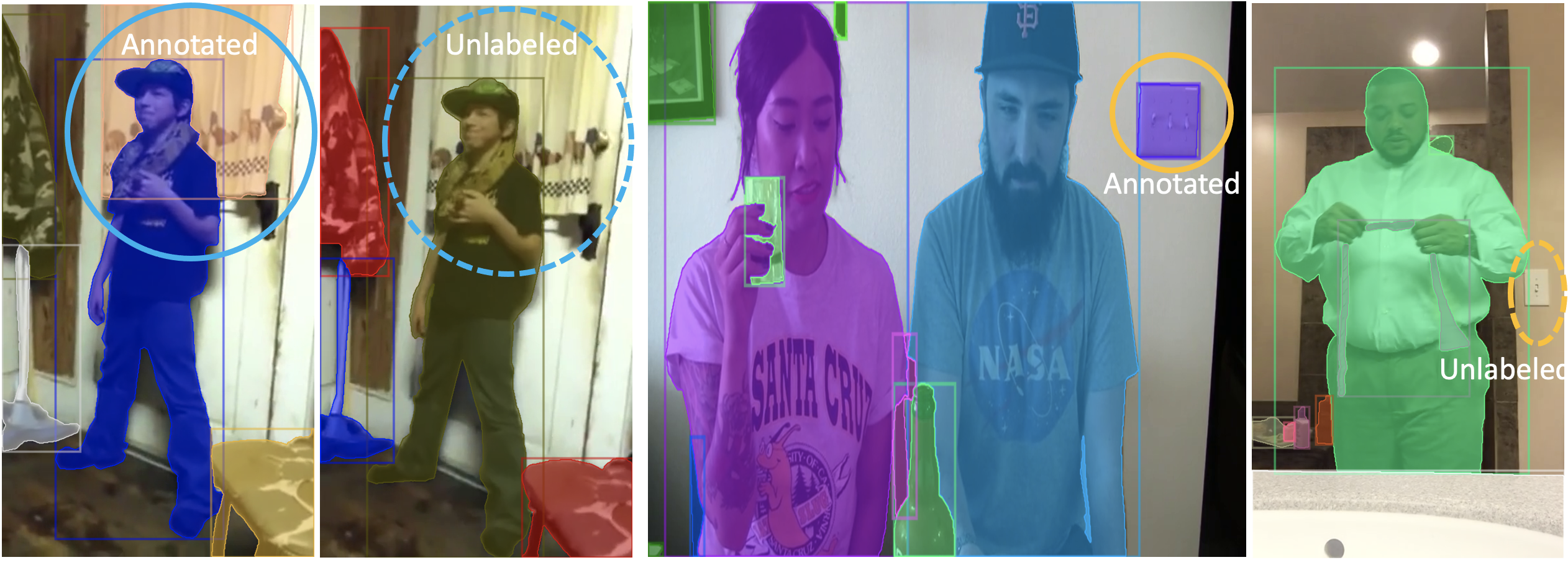}
		\includegraphics[width=1\linewidth]{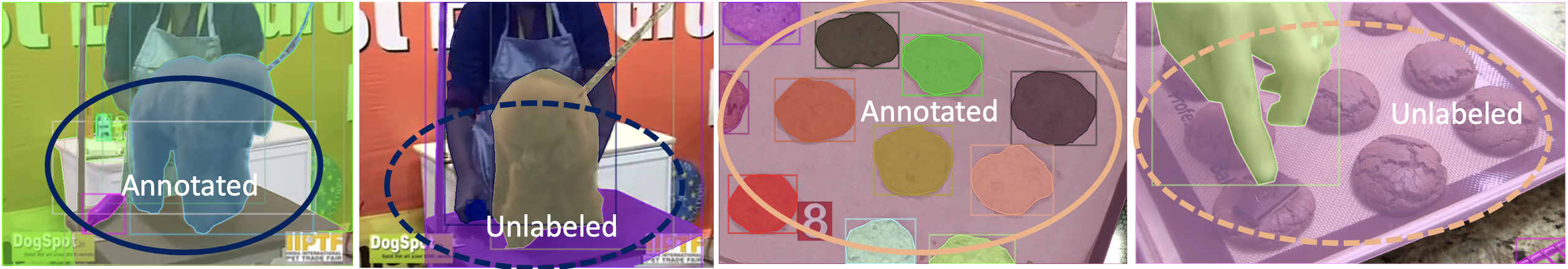}
	\end{center}
	\caption{ Visualizations of UVO annotations. It is notable that the same class of object may be labeled as an instance or as background in different images. (as shown in the area highlighted by the ellipse). This inconsistency of annotations pose a great challenge to the algorithms.}
	\label{fig:fig3}
\end{figure*}

\begin{figure*}[]
	\begin{center}
		\includegraphics[width=1\linewidth]{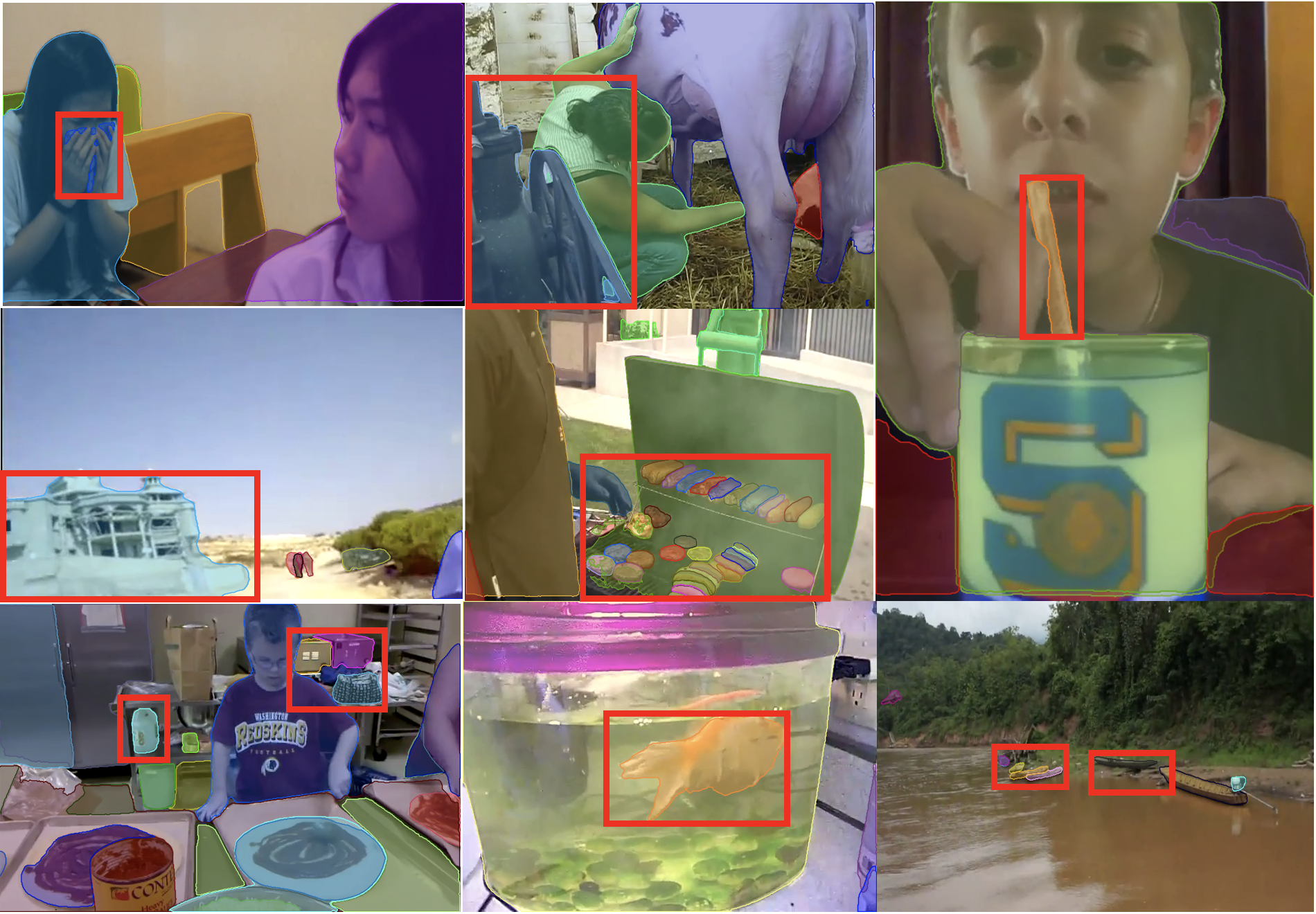}
	    \includegraphics[width=1\linewidth]{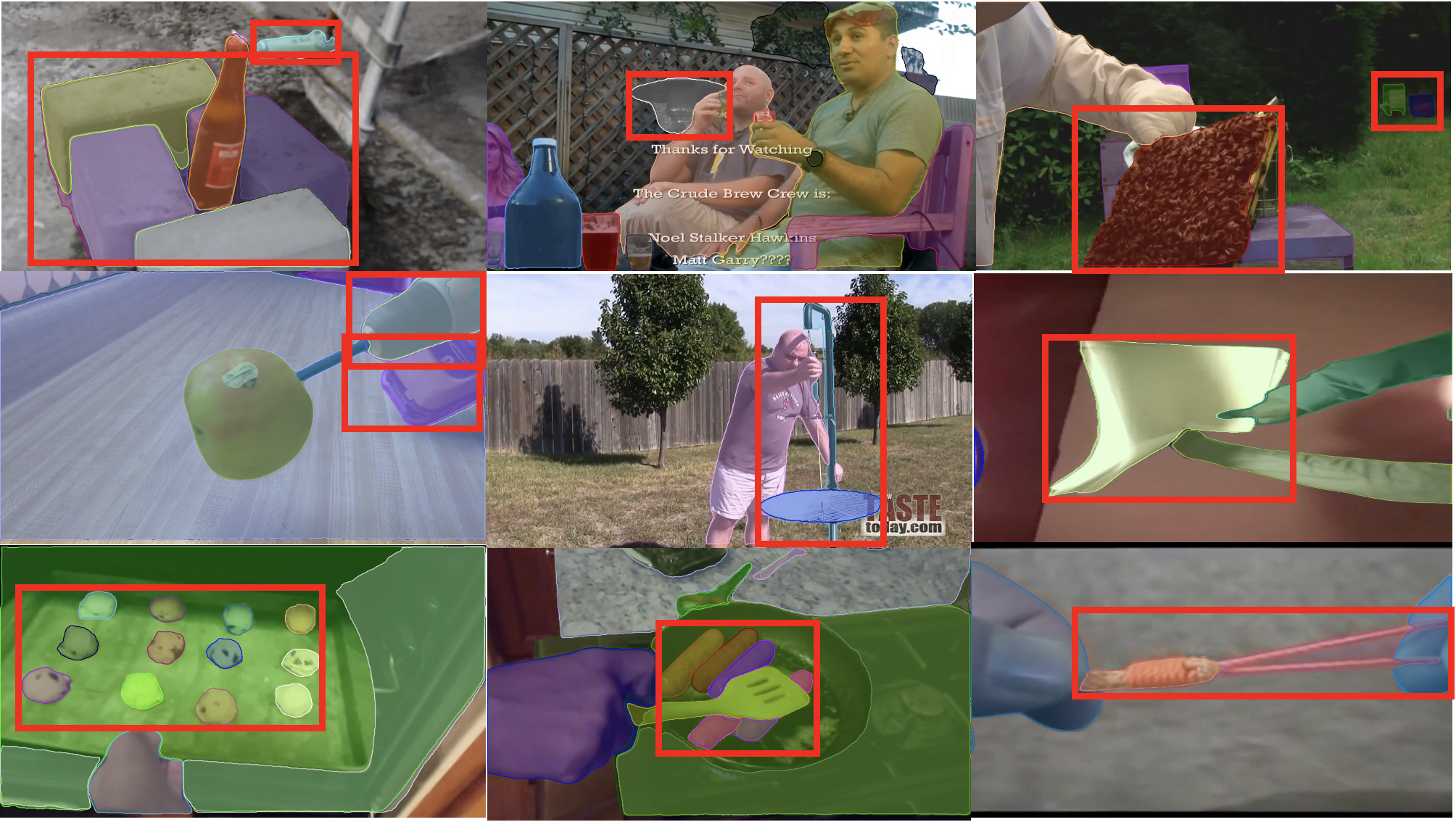}
	\end{center}
	\caption{  Visualizations results of our proposed SOIS in UVO dataset.  SOIS can discover many novel objects, as shown in regions in \textcolor{red}{red boxes}.}
	\label{fig:fig4}
\end{figure*}
% \begin{figure*}[t]
% 	\begin{center}
% 		\includegraphics[width=1\linewidth]{fig/15.png}
% 	    \includegraphics[width=1\linewidth]{fig/16.png}
% 	\end{center}
% 	\caption{  Visualizations results of our proposed SOIS in UVO dataset.  SOIS can discover many novel objects. as shown in regions in \textcolor{red}{red boxes}.}
% 	\label{fig:fig4}
% \end{figure*}
\subsection{Visualization of annotations and our results on UVO dataset}
 Unlike in closed-world instance segmentation, where the object categories have been clearly defined, instance definition in OWIS is much more ambiguous and harder for annotators to follow. Inevitably, the instance annotation could become inconsistent across images, as shown in Figure \ref{fig:fig2}. Our method is motivated by this observation that the instance annotation in the existing datasets is very noisy. Our solution to this issue is to introduce a self-correcting mechanism to combat erroneous annotations, which provides additional guidance to both prediction tasks when the noisy annotations fail to provide correct supervision. The visualization results in Figure \ref{fig:fig3} 
 %and \ref{fig:fig4} 
 demonstrate that our proposed SOIS can segment many novel objects that have not been unseen in the training set.

%%%%%%%%%%%%%%%%%%%%%%%%%%%%%%%%%%%%%%%%%%%%%%%%%%%%%%%%%%%%
\bibliographystyle{ieeetr} 
\bibliography{ref} 
\end{document}